\documentclass[10pt,twocolumn,letterpaper]{article}

\usepackage[cvpr]{cvpr}      
\definecolor{cvprblue}{rgb}{0.21,0.49,0.74}
\usepackage[pagebackref,breaklinks,colorlinks,allcolors=cvprblue]{hyperref}
\usepackage[table]{xcolor}
\usepackage{booktabs}
\usepackage{pifont}         
\usepackage{multirow}
\usepackage{booktabs}
\usepackage{multirow}
\usepackage{makecell}
\usepackage{pifont}
\usepackage{array}

\usepackage{tcolorbox}
\tcbuselibrary{skins,breakable}
\newcommand{\ignore}[1]{}

\makeatletter
\renewcommand\paragraph{\@startsection{paragraph}{4}{\z@}%
  {0.5ex \@plus 0.2ex \@minus .2ex}%
  {-1em}%
  {\normalfont\normalsize\bfseries}}%
\makeatother

\def\paperID{*****} 
\def\confName{CVPR}
\def\confYear{2026}

\newcommand{\lianhui}[1]{\textcolor{red}{[Lianhui: #1]}}

\newcommand{\deliverybench}{\textsc{DeliveryBench}\xspace}

\title{DeliveryBench: Can Agents Earn Profit in Real World?}

\author{%
  Lingjun Mao$^{1}$ \quad
  Jiawei Ren$^{1}$ \quad
  Kun Zhou$^{1}$ \quad
  Jixuan Chen$^{1}$ \quad
  Ziqiao Ma$^{2}$ \quad
  Lianhui Qin$^{1}$ \\
  $^{1}$University of California, San Diego \quad
  $^{2}$University of Michigan \\
  \texttt{lingjun@ucsd.edu}
}

\begin{document}
\maketitle
\begin{abstract}
LLMs and VLMs are increasingly deployed as embodied agents, yet existing benchmarks largely revolve around simple short-term tasks and struggle to capture rich realistic constraints that shape real-world decision making. To close this gap, we propose \deliverybench, a city-scale embodied benchmark grounded in the real-world profession of food delivery. Food couriers naturally operate under long-horizon objectives (maximizing net profit over hours) while managing diverse constraints, \eg delivery deadline, transportation expense, vehicle battery, and necessary interactions with other couriers and customers.
\deliverybench instantiates this setting in procedurally generated 3D cities with diverse road networks, buildings, functional locations, 
transportation modes, and realistic resource dynamics, enabling systematic evaluation of constraint-aware, long-horizon planning. We benchmark a range of VLM-based agents across nine cities and compare them with human players.
Our results reveal a substantial performance gap to humans, and find that these agents are short-sighted and frequently break basic commonsense constraints. Additionally, we observe distinct personalities across models (\eg adventurous GPT-5 vs. conservative Claude), highlighting both the brittleness and the diversity of current VLM-based embodied agents in realistic, constraint-dense environments. Our code, data, and benchmark are available at \url{https://deliverybench.github.io}.

\end{abstract}

\section{Introduction}

\definecolor{groupfilllight}{gray}{0.96}   
\DeclareRobustCommand{\cmark}{\ding{51}}   
\DeclareRobustCommand{\xmark}{\ding{55}}   

\newcommand{\GroupRowEight}[1]{%
  \rowcolor{groupfilllight}\multicolumn{8}{l}{\textsc{#1}}\\[-0.2ex]
}

\begin{table*}[t]
\caption{\textbf{Comparison of major embodied benchmarks.}
Benchmarks are compared across sequence length per episode and six constraint dimensions, with \deliverybench featuring longer horizons and more comprehensive multidimensional constraints (see  Section~\ref{sec:constraints}).}
\vspace{-5pt}

\label{tab:comparison}
\centering

\begin{minipage}[t]{0.663\textwidth}
  \vspace{0pt}
  \footnotesize
  \setlength{\tabcolsep}{4.0pt}
  \renewcommand{\arraystretch}{1.12}
  \resizebox{\linewidth}{!}{%
  \begin{tabular}{cccccccc}
  \toprule
\multirow{2}{*}{\textbf{Benchmark}} &
\multirow{2}{*}{\makecell{\textbf{Sequence Length}\\\textbf{(action steps)}}} &
\multicolumn{6}{c}{\textbf{— Task Constraints —}} \\
\cmidrule(lr){3-8}
& & \textbf{Spatial} & \textbf{Time} & \textbf{Resource} &
\textbf{Physical} & \textbf{Economic} & \textbf{Social} \\
  \midrule

  BEHAVIOR~\citep{pan2024large}       & —        & \xmark & \xmark & \xmark & \cmark & \xmark & \xmark \\
  ManiSkill2~\citep{gu2023maniskill2} & —        & \xmark & \xmark & \xmark & \cmark & \xmark & \xmark \\
  CookBench~\citep{cai2025cookbench}  & $>100$   & \cmark & \cmark & \xmark & \cmark & \xmark & \xmark \\
  ALFRED~\citep{shridhar2020alfred}       & $\sim$12 & \cmark & \xmark & \xmark & \cmark & \xmark & \xmark \\
  ReALFRED~\citep{kim2024realfred}        & $\sim$12 & \cmark & \xmark & \xmark & \cmark & \xmark & \xmark \\
  EB-ALFRED~\citep{yang2025embodiedbench} & $\sim$12 & \cmark & \xmark & \xmark & \cmark & \xmark & \xmark \\
  ALFWorld~\citep{shridhar2020alfworld}   & $\sim$6  & \cmark & \xmark & \xmark & \xmark & \xmark & \xmark \\
  VirtualHome~\citep{puig2018virtualhome} & $\sim$9  & \cmark & \xmark & \xmark & \cmark & \xmark & \xmark \\
  ET-Plan-bench~\citep{zhang2024plan}     & $<$20    & \cmark & \xmark & \xmark & \cmark & \xmark & \xmark \\
  EmbRACE-3K~\citep{lin2025embrace3kembodiedreasoningaction}
                                          & $\sim$10 & \cmark & \xmark & \xmark & \cmark & \xmark & \xmark \\
  TEACh~\citep{padmakumar2022teach}       & —        & \cmark & \xmark & \xmark & \cmark & \xmark & \cmark \\
  ProcTHOR~\citep{deitke2022️}            & —        & \cmark & \xmark & \xmark & \cmark & \xmark & \xmark \\
  TaPA~\citep{wu2023embodied}             & $\sim$25 & \cmark & \xmark & \xmark & \cmark & \xmark & \xmark \\

  \textbf{\deliverybench}                   & \textbf{$>100$}
                                            & \textbf{\cmark} & \textbf{\cmark} & \textbf{\cmark}
                                            & \textbf{\cmark} & \textbf{\cmark} & \textbf{\cmark} \\
  \bottomrule
  \end{tabular}%
  }
\end{minipage}
\quad
\quad
\begin{minipage}[t]{0.29\textwidth}
  \vspace{-3.0pt}
  \centering
  \includegraphics[width=\linewidth]{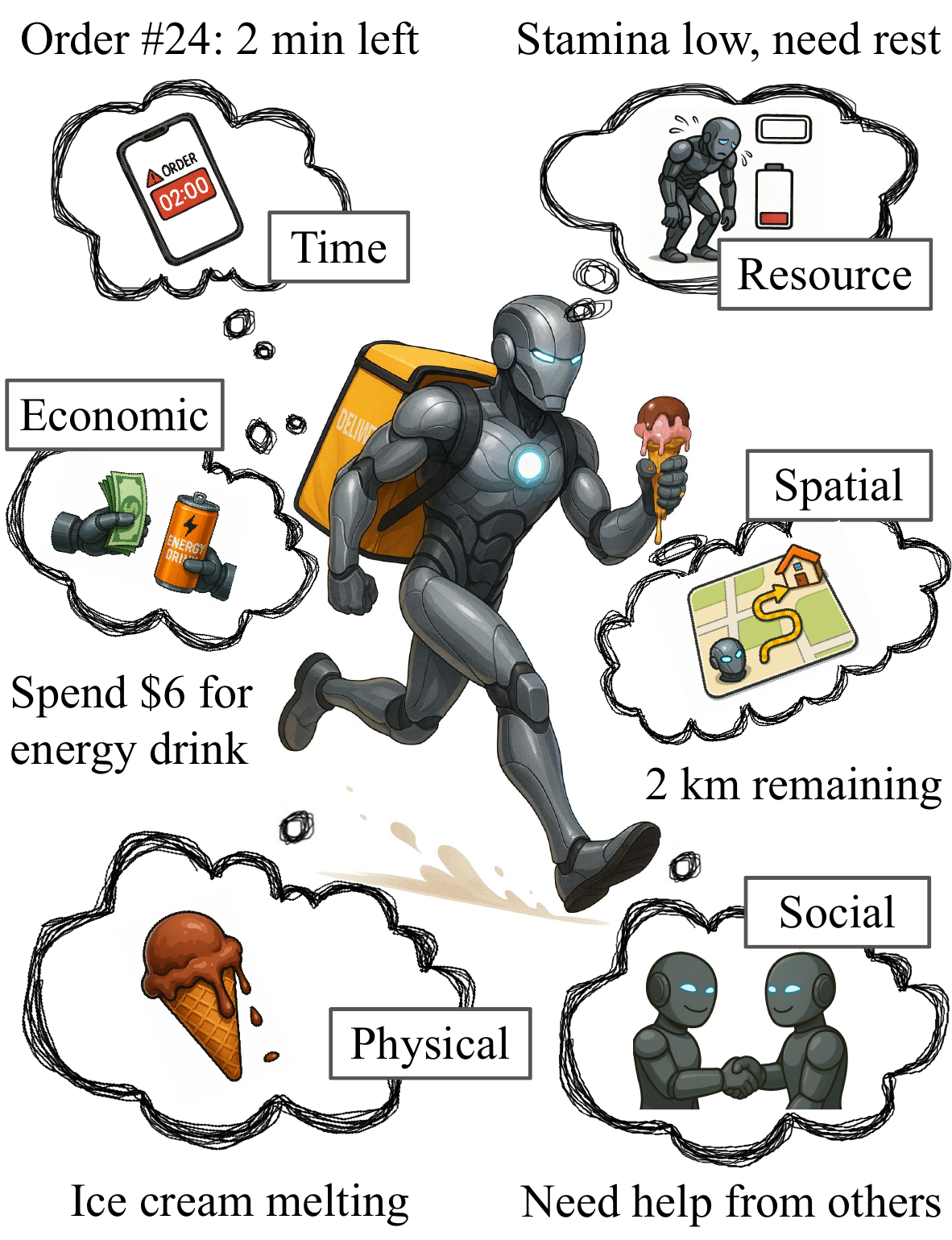}
\end{minipage}

\vspace{-15pt}
\end{table*}

Large language models (LLMs) and vision-language models (VLMs) have exhibited strong abilities in solving diverse real-world problems, such as mathematics \citep{luo2025large, wang2025mathcoder} and programming \citep{robeyns2025self, claude2025}. Building on these advances, recent research has begun exploring embodied agents that can perceive, reason, and act in physical environments \citep{liu2024visualagentbench, hong2025embodied, kim2025beyond, islam2023eqa, li2024embodied}. Looking ahead, humans increasingly envision AI agents that may one day operate autonomously in the real world, helping with household tasks, participating in scientific discovery, or even earning income on our behalf. To move toward this vision, the community has developed a series of embodied-agent planning benchmarks that approximate real-world challenges through simulated environments, including 3D simulators \citep{yang2025embodiedbench, cheng2025embodiedeval, zhong2025unrealzoo} and open-world games such as Minecraft \citep{white2025collaborating, long2024teamcraft}. By defining grounded tasks and modeling realistic constraints, these platforms help evaluate emerging agent abilities and provide data to guide future system design or model training.

A core capability for autonomous agents operating in the real world is to \textbf{earn profit and sustain themselves economically}. Beyond completing isolated tasks, a truly capable agent should be able to survive, adapt, and even develop a long-term career, navigating decisions that balance cost, benefit, and risk in the real world. Building and evaluating such agents requires environments that faithfully reflect the complexity of everyday life, where decisions unfold over long horizons, and outcomes depend on interacting physical, economic, resource, and social factors. To study it, a realistic benchmark should not only support embodied perception and action, but also model the incentives, constraints, and trade-offs that determine whether an agent can accumulate profit and operate sustainably. However, as shown in Table~\ref{tab:comparison}, existing benchmarks fall short of this goal. They either overemphasize short-horizon subtasks (\eg navigation, pickup-and-drop) or fail to encode the nontrivial constraints that shape real decision-making.

In this paper, we aim to introduce a realistic embodied-agent benchmark that demands long-horizon planning while adhering to multiple real-world constraints. To minimize the gap between simulation and reality, such a benchmark must be grounded in tasks that (i) \emph{truly exist in the real world}, (ii) \emph{naturally involve long-term objectives}, and (iii) \emph{require to simultaneously manage diverse constraints}.
After surveying a variety of real-world careers, we find that food delivery provides an ideal testbed. A delivery courier operating in a city must carefully sequence routes using appropriate transportation, interleave supportive actions (\eg recharging an e-scooter or purchasing tickets), and collaborate with others when needed, all to maximize completed orders and net earnings. An example is shown in Figure~\ref{tab:comparison}.

We develop \deliverybench, a city-scale benchmark that evaluates embodied agents under physically and socially grounded delivery scenarios. Agents act as autonomous couriers navigating procedurally generated cities to maximize long-term profit. To capture the open-ended nature of real-world operations, \deliverybench features dynamic, interactive environments populated with diverse points of interest (POIs) and multiple modes of transportation, going beyond prior urban simulators \citep{embodiedcity2024, hong2025embodied, wu2024metaurban} that primarily offer static visual scenes. As deliveries unfold across multiple in-game hours, agents must manage resources (\eg stamina depletion), adapt to changing conditions, and strategically balance efficiency, timing, and cost. When multiple agents coexist, they further encounter social dynamics such as competition and collaboration. By jointly modeling economic, physical, and social dynamics within a unified embodied environment, \deliverybench provides a realistic and action-driven setting to test whether VLM-based agents can make and execute plans that genuinely improve financial outcomes.

Using \deliverybench, we conduct extensive experiments on (i) a diverse set of state-of-the-art VLMs, (ii) under both single-agent and multi-agent settings, and (iii) across nine cities with distinct geographic layouts. Our results reveal several findings. Frontier VLM-based agents lag far behind human players, struggling with long-horizon, constraint-aware decision making and frequently making naïve mistakes (\eg forgetting to recharge an e-scooter). Multi-agent performance does not scale with team size and typically peaks with two-agent teams, suggesting coordination challenges. Context engineering on larger models yields significant gains in improving the earned profit.
Finally, different VLMs exhibit distinct behavioral styles—GPT-5 appears adventurous, Claude more conservative, and Gemini comparatively careless.

\section{Related Works}


\paragraph{VLM-based Embodied Agent.} Recent advances in VLMs~\citep{openai2025gpt5card, 2025claude3.7sonnet, comanici2025gemini} and large-scale manipulation datasets~\citep{o2024open, bu2025agibot} have driven the development of embodied agents~\citep{zitkovich2023rt, driess2023palm, yang2025agentic} that translate language instructions into grounded visual understanding and executable actions. Although these models have shown strong performance on short-horizon tasks, they still struggle with complex long-horizon scenarios, motivating the emergence of new agentic-workflow designs~\citep{wang2023voyager, mu2023embodiedgpt} and training-based approaches~\citep{zitkovich2023rt, driess2023palm, yang2025lohovla} in embodied settings. Agentic workflows aim to improve model adaptivity by incorporating mechanisms such as explicit memory~\citep{lei2025clea}, reflection~\citep{huang2022inner, yang2025lohovla}, and feedback-driven correction~\citep{yang2025guiding, kumar2024open}. In contrast, training-based approaches emphasize end-to-end~\citep{intelligence2504pi0} or distilled learning~\citep{sumers2023distilling} frameworks that unify perception, reasoning, and control. Yet, it remains unclear how well these embodied agent designs perform when faced with tasks that truly reflect the long-horizon nature and complexity of real-world settings.

\paragraph{Embodied Agent Benchmarks.} Existing embodied benchmarks vary widely in abstraction level and planning horizon.
Low-level control benchmarks such as BEHAVIOR~\citep{srivastava2021behavior}, iGibson~\citep{shen2021igibson},
SAPIEN~\citep{xiang2020sapien}, and ManiSkill2~\citep{gu2023maniskill2} emphasize fine-grained
motor control and physical realism, requiring precise actuator adjustment and object manipulation.
These environments rely on high-fidelity physics engines (e.g., MuJoCo~\citep{todorov2012mujoco},
PyBullet~\citep{coumans2016pybullet}) to simulate realistic dynamics and evaluate action-level precision. By contrast, long-horizon embodied benchmarks such as ALFRED~\citep{shridhar2020alfred},
ReALFRED~\citep{kim2024realfred}, and TEACh~\citep{padmakumar2022teach} emphasize multi-step
instruction following (typically 10–30 steps) and structured task planning. Later extensions (e.g., ProcTHOR~\citep{deitke2022️},
EmbRACE-3K~\citep{lin2025embrace3kembodiedreasoningaction}) expand scene diversity and interaction
complexity, while others such as VirtualHome~\citep{puig2018virtualhome}, ALFWorld~\citep{shridhar2020alfworld},
and ET-Plan-bench~\citep{zhang2024plan} abstract tasks into programs or textual plans to probe reasoning and decomposition abilities. However, existing benchmarks often overlook multidimensional constraints (e.g., economic, resource, or social) and still fall short of truly open-ended, long-horizon decision-making. We introduce \deliverybench\ to address these gaps.

\begin{figure*}[!h]
    \centering
    \includegraphics[width=0.95\linewidth]{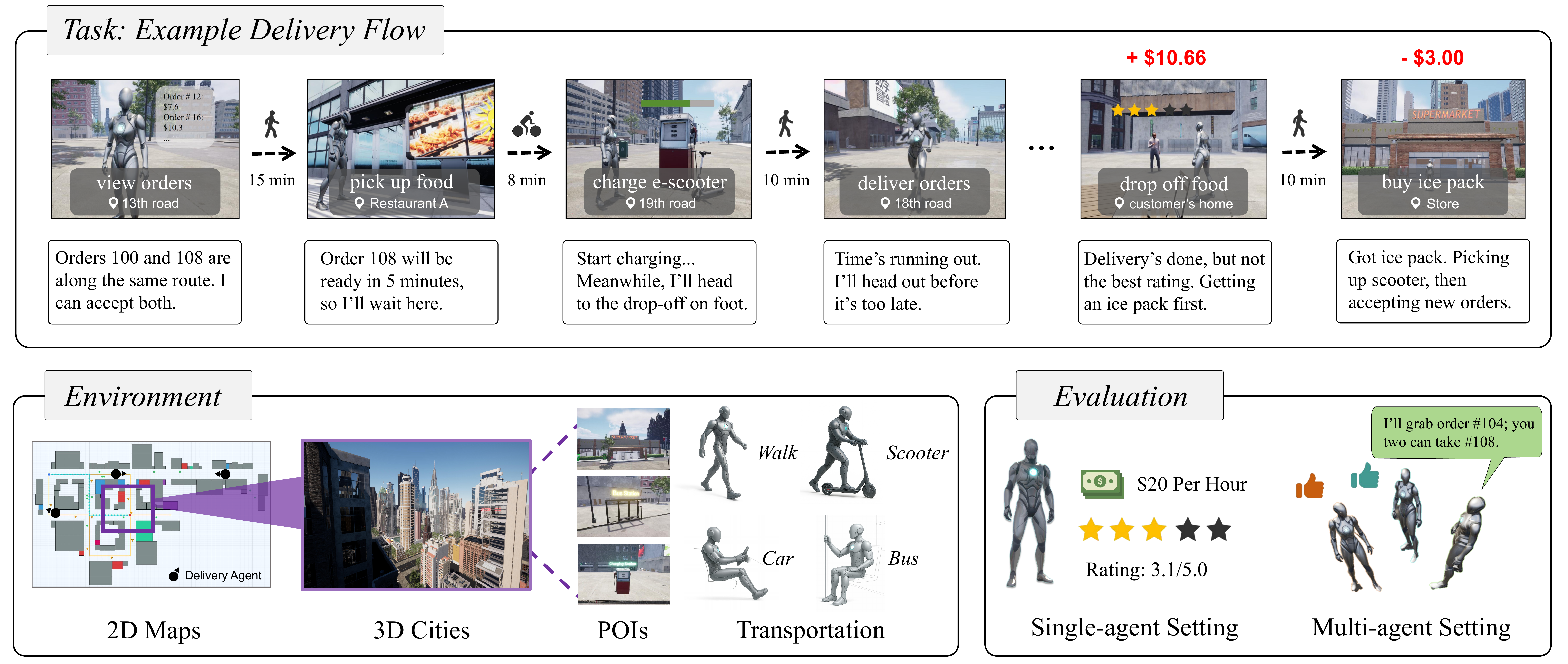}
    \vspace{-3pt}
    \caption{Overview of the \deliverybench environment. The process consists of both core delivery actions (e.g., viewing, accepting, picking up, and delivering orders) and supporting actions (e.g., recharging e-scooters, purchasing items) that assist sustained delivery.}
    \label{fig:DeliveryBench}
    \vspace{-10pt}
\end{figure*}

\section{DeliveryBench}
\label{sec:deliverybench}

In this section, we present our \deliverybench, a long-horizon planning benchmark for evaluating VLM-based embodied agents under realistic, constraint-rich settings. \deliverybench integrates heterogeneous task objectives, realistic multifaceted constraints, and diverse evaluation dimensions.
An overview is illustrated in Figure~\ref{fig:DeliveryBench}.

\subsection{Profit-Earning Task}
We center our benchmark on the food-delivery scenario, where an agent works in a virtual city and aims to maximize net profit by continuously completing delivery orders. 


\subsubsection{Task Formulation}
The delivery task is formalized as a \emph{long-horizon constrained optimization problem}, where a VLM-based agent as a courier seeks to maximize \emph{net profit} over an operational horizon $T$ (\eg two virtual hours).
To do so, the agent must plan and execute a sequence of delivery and supportive tasks while respecting diverse real-world constraints.

\paragraph{Long-term Profit Target.}
The agent earns income from customer orders in two forms:
(i) a base salary $E_{\text{base}}$ upon successful delivery; and
(ii) rating-based rewards $E_{\text{rating}}$, determined by factors such as delivery punctuality, freshness, and special instructions (\eg face-to-face delivery).
Meanwhile, operational costs (C) arise from purchasing items or services (\eg recharging, vehicle rental).
The total income and net profit are therefore
\begin{equation}
E = E_{\text{base}} + E_{\text{rating}}, \qquad P = E - C.
\end{equation}

\vspace{-5pt}
\paragraph{Constrained Decision Making.}
At each step, the agent receives an observation $O_t$ and selects an action $a_t = \pi_i(O_t)$ via policy $\pi_i$.
The goal is to obtain an optimal policy $\pi_i^\star$ that maximizes expected net profit while satisfying all constraints $\mathcal{C}$.
Let $\Pi_{\mathcal{C}}$ be the set of feasible policies whose induced trajectories obey all $c \in \mathcal{C}$.
Formally,
\begin{equation}
\pi_i^\star \in \arg\max_{\pi_i \in \Pi_{\mathcal{C}}} \mathbb{E}_{\pi_i}\!\left[P\right].
\end{equation}
To achieve this objective, the agent must coordinate both delivery-related tasks that directly contribute to revenue (\eg selecting, fulfilling orders, or managing freshness decay) and supportive tasks that indirectly maintain operational feasibility (\eg recharging, resting, purchasing supplies, or renting vehicles).


\subsubsection{Test Environment}
To support realistic and versatile task execution, we simulate a high-fidelity 3D urban environment featuring diverse city layouts, interactive points of interest (POIs), multiple transportation modes, and rich physical dynamics.

\paragraph{Simulated 3D City.}
Based on SimWorld~\citep{ren2025simworld}’s procedural generator, in \deliverybench, we simulate different scales of 3D city layouts inside Unreal Engine.
Each city contains realistic buildings, roads, humans, and other objects, where the complete action trajectory of the agent can be logged and visualized to the user for monitoring and evaluation.
Besides, the realistic weather control, physics simulation, and other features inside Unreal Engine, support us to flexibly vary the environments and ensure the reality.

\paragraph{Interactive Infrastructure and POIs.}
Across all cities, buildings are sampled as POIs with equal probability, including restaurants, customer homes, convenience stores, car rentals and rest areas.
Infrastructure such as bus stops and charging stations is placed along the road network. 
When an agent arrives at these these POIs and infrastructures, it can trigger context-specific actions (\eg picking up food, recharging vehicles, renting cars, or resting).

\paragraph{Transportation, Navigation, and Physics.}
The environment supports multiple transportation modes (\eg walking, e-scooters, cars, and public transit), with different speed, cost, and stamina profiles. 
Because current models struggle with low-level 3D navigation~\citep{ramrakhya2022habitat, song2025towards}, we provide a waypoint-based system that follows shortest paths while still exposing motion control. Physical dynamics (\eg temperature, collisions, odor diffusion) further affect food quality during transit, requiring agents to adapt routing and mode choices to preserve freshness.

\subsection{Multifaceted Realistic Constraints}
\label{sec:constraints}

\deliverybench\ is designed to expose agents to the types of constraints that structure real-world decision making. 
As summarized in Table~\ref{tab:comparison}, we categorize these constraints into six major types: 
\emph{spatial}, \emph{time}, \emph{resource}, \emph{physical}, \emph{economic}, and \emph{social}. 
Each type governs what actions are feasible and how desirable different plans are, and together they induce a rich, tightly coupled planning landscape.

\begin{itemize}[leftmargin=10pt]

\item \textbf{Spatial constraints:}
Spatial constraints specify \emph{where} actions can be executed. 
Certain operations are only valid at designated POIs: for instance, order pickup must occur at the associated restaurant, and recharging is only possible at charging stations. 
The agent must therefore navigate the city and visit appropriate POIs in a coherent sequence to complete deliveries and supportive tasks.

\item \textbf{Time constraints:}
Time constraints restrict \emph{when} tasks can be performed. 
Each task is associated with a feasible time window, and some tasks must follow others in a fixed order (\eg a delivery must happen after the corresponding pickup). 
When windows overlap without ordering requirements, the agent can interleave tasks to improve efficiency, such as delivering an existing order while waiting for a new meal to be prepared.
Some tasks also have deadlines: late deliveries reduce income, and the overall episode is limited by a maximum working duration, forcing the agent to use its time budget carefully.

\item \textbf{Resource constraints:}
Agents must manage consumable resources such as stamina, vehicle battery, and cash to stay operational.
Depleting any resource impairs related abilities (\eg cannot ride a e-scooter without recharging).
To stay self-sustained, the agent needs to schedule supportive actions such as resting, recharging, or purchasing consumables, and can sometimes convert one resource into another, \eg spending cash to restore stamina.

\item \textbf{Physical constraints:}
Physical constraints capture how environmental dynamics affect delivery outcomes.
Temperature, motion, and collisions all influence food condition (\eg ice cream melts, fragile items can be damaged). 
As a result, route planning and transport mode must consider not only distance and time but also the fragility and perishability of delivered items.

\item \textbf{Economic constraints:}
Economic constraints arise from the balance between income and cost.
Agents can earn money from base pay and rating-based bonuses, but incur expenses for actions such as recharging vehicles, renting cars, or buying supplies. 
Some of these expenses can be viewed as investments in long-term gains, requiring agents to balance immediate costs against future benefits.

\item \textbf{Social constraints:}
In multi-agent settings, multiple couriers operate in the same city, introducing additional constraints from \emph{collaboration} and \emph{competition}. 
Agents may coordinate implicitly or explicitly, for example by serving different regions or handing off orders and resources, but they also compete for scarce opportunities such as high-value orders and nearby charging spots. 

\end{itemize}

\subsection{Benchmark Construction}
In this part, we describe how we build \deliverybench, outline the task setup for both single- and multi-agent settings, and introduce metrics to evaluate the multi-dimensional capabilities of VLM-based agents.

\subsubsection{Task Setup}

\paragraph{Multi-level Tasks Creation.}
We evaluate agents on nine procedurally generated city maps covering three difficulty levels: 
\emph{small} (11–15 roads), \emph{medium} (16–25 roads), and \emph{large} (26–30 roads). 
Each environment maintains an order pool with a fixed number of active delivery orders, which is continuously replenished as orders are accepted. 
For each order, the system randomly samples a restaurant (pickup location) and a residential building (dropoff location); the delivery wage and time limit are then computed from the travel distance with slight stochastic perturbations for variability. 
We maintain a certain percentage of orders contain special customer requirements (\eg face-to-face delivery), and violations incur penalties. 
Each episode terminates when the agent reaches either the lifetime or API calls budget.

\paragraph{Agent State Management.}
At the beginning of each episode, agents are spawned at a designated starting location in the city. 
All agents share the same embodiment, camera configuration, and base movement speed. 
Their initial states are the same, with an initial value of the stamina, balance, battery level and other related features.
As agents act, stamina and battery levels decrease according to their activities. 
At the end of each episode, we log the complete interaction trajectory, income, and expenses, which form the basis for our evaluation metrics.

\subsubsection{Single- and Multi-agent Settings}

\paragraph{Single-agent regime.}
In the single-agent setting, one agent operates as the sole courier in each city. 
This regime isolates individual planning, reasoning, and constraint-handling ability without interference from other agents. 
Each agent is evaluated on all nine maps under the same task-generation process and episode termination criteria, with results averaged over multiple separate runs.

\paragraph{Multi-agent regime.}
In the multi-agent setting, we deploy eight instances of the same agent in a shared environment to study competition and cooperation. 
All agents draw from a global order pool and share infrastructure such as charging stations, producing competition for high-value orders and scarce resources.
To control the degree of cooperation, we group them into different team structures: 
\(8\times 1\) (eight independent agents, purely competitive), 
\(4\times 2\) (four cooperating pairs), 
\(2\times 4\) (two groups of four), and 
\(1\times 8\) (a single fully cooperative team). 
Within each group, agents can communicate and respond to help requests, enabling behaviors such as handing off orders and recharging a teammate’s e-scooter. 
This design probes how social structure and team size affect performance and interaction patterns.

\subsubsection{Evaluation Metrics}

\paragraph{Global profit.}
Our primary performance metric is the hourly net profit $\bar{P}$ achieved in a 2-hour virtual episode. 
We report $\bar{P}$ aggregated over episodes as the main indicator.

\paragraph{Fine-grained Capability Analysis.}
To diagnose where agents succeed or fail, we further evaluate model behavior along following three capability dimensions, and more details about the evaluation metrics are in Appendix~\ref{appendix:metrics}.

\begin{itemize}[leftmargin=10pt]
    \item \textbf{High-level planning.}
    We measure time-sensitive long-term planning via order-selection quality, on-time delivery rate, time efficiency (effective delivery time including parallel orders, normalized by episode time), and active time ratio (fraction of time spent on purposeful actions rather than idling or being incapacitated).
    
    \item \textbf{Resource management.}
    We assess self-sustaining behavior using hourly stamina consumption, interruption count (\eg stops due to resource depletion), and proactive prevention ratio (how often agents replenish critical resources before they run out).
    
    \item \textbf{Physical/environmental adaptation.}
    We evaluate how well agents handle implicit physical and environmental constraints using violation rate (fraction of orders with constraint violations), food-quality rating, and customer rating (both on a 0–5 scale). 
    These metrics capture whether agents can handle realistic constraints.
\end{itemize}


\vspace{-10pt}
\begin{figure}[!h]
    \centering
    \includegraphics[width=0.9\linewidth]{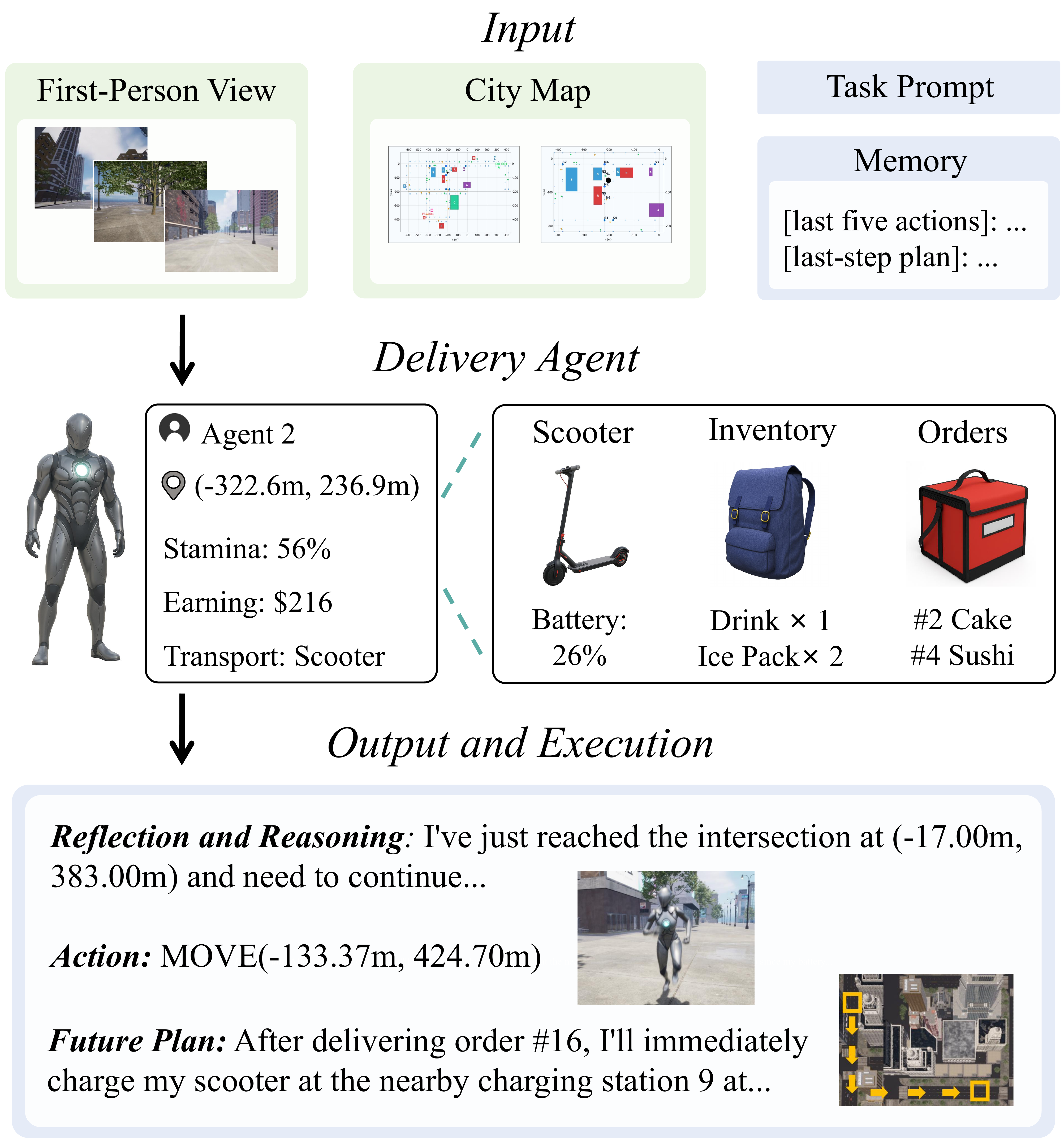}
    \vspace{-5pt}
    \caption{Overview of the agent’s perception–planning–execution loop in \deliverybench.}
    \label{fig:agent-io}
    \vspace{-15pt}
\end{figure}
\section{Agent Design}

Each agent follows an perception–planning–execution loop and operates as a high-level planner over a rich embodied environment. 
At each timestep $t$, the agent perceives the city, reasons about its current tasks and constraints, and selects an action to update its trajectory and long-term plan. The framework is illustrated in Figure~\ref{fig:agent-io}.

\paragraph{Observation Space.}
The observation space aggregates multiple complementary views of the city and the agent’s operational status. 
A \emph{global map} $o^{\text{global}}_t$ shows the full city layout, including the agent’s location and major points of interest (POIs); 
a \emph{local map} $o^{\text{local}}_t$ provides finer-grained details of the nearby area; 
and a \emph{first-person view} (FPV) $o^{\text{fpv}}_t$ renders the agent’s embodied perspective, capturing streets, buildings, and surrounding objects. 
In addition, the agent can query \emph{auxiliary information} $o^{\text{aux}}_t$ via explicit actions, such as checking current orders, inventory, or public transport schedules. 
The full observation at time $t$ is thus
\[
O_t = \{\, o^{\text{global}}_t,\; o^{\text{local}}_t,\; o^{\text{fpv}}_t,\; o^{\text{aux}}_t \,\}.
\]

\paragraph{Action Space.}
The action space in \deliverybench\ supports both high-level decision making and fine-grained embodied control, denoted as $\mathcal{A}$. We provide its full details in Appendix~\ref{appendix:actionspace}.
High-level actions allow the agent to delegate complex procedures to the simulator; for example, \texttt{MOVE\_TO} takes a target coordinate (or POI) and triggers automatic path planning and navigation along the road network. 
Low-level actions provide direct control over movement and orientation, such as \texttt{STEP\_FORWARD} or \texttt{TURN\_AROUND}. 
Interaction actions enable the agent to manipulate the environment and manage resources, including picking up or dropping off orders, purchasing or using tools (\eg batteries), and recharging or renting vehicles.

\paragraph{Planning Pipeline.}
To model decision making over long horizons, we adopt a lightweight planning pipeline. 
At timestep $t$, the agent receives the current observation $O_t$ and maintains a short-term memory $M_t = \{a_{t-k:t-1}\}$ of its past $k$ actions. 
It also conditions on the previous plan $P_t$, produced at timestep $t{-}1$, and the most recent failure signal $F_{t-1}$, which indicates whether the last action or plan did not succeed as intended. 
The policy $\pi_\theta$ then outputs both the current action $a_t \in \mathcal{A}$ and an updated plan $P_{t+1}$:
\[
(a_t, P_{t+1}) = \pi_\theta(O_t, M_t, P_t, F_{t-1}).
\]
Through this iterative update mechanism, the agent can continuously refine its future plan while reacting to new observations and failures in the environment, enabling more stable and adaptive behavior over long time horizons.

\begin{table*}[h]
\centering
\caption{
Global performance of different models across city sizes, measured by average hourly net profit (\$/h), with detailed breakdown into base earnings ($E_{\text{base}}$), rating-based bonuses or penalties ($E_{\text{rating}}$), and expenses ($C$).
}
\label{tab:global_performance}
\vspace{-5pt}

\footnotesize
\setlength{\tabcolsep}{4.5pt}     
\renewcommand{\arraystretch}{1.05}

\begin{tabular*}{\textwidth}{@{\extracolsep{\fill}} l|cccc|cccc|cccc}
\toprule
\multirow{2}{*}{\textbf{Model}} &
\multicolumn{4}{c|}{\textbf{Small City}} &
\multicolumn{4}{c|}{\textbf{Medium City}} &
\multicolumn{4}{c}{\textbf{Large City}} \\
\cmidrule(lr){2-5} \cmidrule(lr){6-9} \cmidrule(l){10-13}
& $\bar{P}$ & $E_{\text{base}}$ & $E_{\text{rating}}$ & $C$ &
$\bar{P}$ & $E_{\text{base}}$ & $E_{\text{rating}}$ & $C$ &
$\bar{P}$ & $E_{\text{base}}$ & $E_{\text{rating}}$ & $C$ \\
\midrule
GPT-5                    & \$27.4 & \$31.1 & \$11.5 & \$15.2 & \$26.5 & \$32.9 & \$7.6 & \$14.0 & \$20.4 & \$25.6 & \$8.3 & \$13.4 \\
GPT-4o                   & \$10.4 & \$23.6 & \$6.8 & \$20.0 & \$13.9 & \$25.4 & \$4.9 & \$16.3 & \$11.9 & \$20.6 & \$5.3 & \$13.9 \\
Claude-3.7-Sonnet        & \textbf{\$31.3} & \$30.1 & \textbf{\$14.8} & \$13.6 & \textbf{\$31.2} & \textbf{\$35.7} & \textbf{\$10.5} & \$14.9 & \textbf{\$25.8} & \textbf{\$30.1} & \textbf{\$13.0} & \$17.2 \\
Gemini-2.5-Flash         & \$30.4 & \textbf{\$34.8} & \$10.7 & \$15.0 & \$29.0 & \$32.3 & \$8.3 & \$11.5 & \$23.9 & \$27.2 & \$9.0 & \$12.3 \\
\midrule
Qwen2.5-VL-72B-Ins       & \$5.4 & \$15.1 & \$3.8 & \$13.5 & \$6.3 & \$15.6 & \$3.3 & \$12.6 & -\$2.7 & \$6.4 & \$1.1 & \$10.3 \\
Qwen2.5-VL-32B-Ins       & \$9.8 & \$15.7 & \$5.5 & \$11.4 & \$4.4 & \$11.5 & \$4.5 & \$11.5 & -\$0.1 & \$8.7 & \$2.3 & \$11.1 \\
LLaMA-3.2-90B-Vision-Ins & \$6.0 & \$9.7 & \$2.0 & \$5.7 & \$2.5 & \$11.6 & \$2.3 & \$11.4 & -\$0.9 & \$7.0 & \$1.3 & \$9.3 \\
\midrule
Human                    & \$63.6 & \$77.8 & \$24.4 & \$38.6 & \$51.5 & \$73.6 & \$12.8 & \$34.9 & \$55.4 & \$74.3 & \$12.8 & \$31.6 \\
\bottomrule
\end{tabular*}
\vspace{-5pt}
\end{table*}

\section{Experiments}

\subsection{Experimental Setup}

\paragraph{Simulation Protocol.}
Our evaluation spans nine procedurally generated city maps, distributed across three difficulty levels.
The order pool maintains 10 active orders, with 40\% containing special customer requirements. 
We fix the weather to sunny with a temperature of 22°C. 
All VLM-based agents start with full stamina, an initial balance of \$100, and an e-scooter at 50\% battery, together with basic insulation to slow food-quality degradation during transit.
Agents continue acting in the virtual world until they reach either a 2-hour lifetime budget or a cap of 300 API calls. The simulation speed is set to three times that of real time. 
To avoid bias from model response latency, we pause each agent’s lifetime clock, order timers, and food dynamics while it is reasoning. Time only advances when actions are executed.
We fix random seeds to ensure identical order generation across runs. Each model is evaluated over eight independent runs per map, reporting average performance.

\paragraph{Baseline Models.}
We test seven representative models: four closed-source models (GPT-5~\citep{openai2025gpt5card}, GPT-4o~\citep{2025gpt4omini}, Claude-3.7-Sonnet~\citep{2025claude3.7sonnet}, and Gemini-2.5-Flash~\citep{comanici2025gemini}) and three open-source models (Qwen2.5-VL-72B~\citep{bai2025qwen2}, Qwen2.5-VL-32B, and LLaMA-3.2-90B-Vision~\citep{meta2024llama32vision}). 
For GPT-5, we use the ``minimal'' reasoning effort setting. 
We fix a temperature of 0 and a maximum completion length of 512 tokens. VLMs are accessed via the OpenRouter\footnote{\url{https://openrouter.ai/}}.

\paragraph{Human Baseline.}
To establish a meaningful reference for single-agent performance, we include a human baseline by recruiting three participants to independently complete the same delivery tasks. Each participant interacts via a custom GUI and follows the same evaluation protocol as the models. Interface details and screenshots are provided in the Appendix~\ref{appendix:humangui}. We also record their delivery trajectories for subsequent supervised fine-tuning experiments.

\subsection{Single-Agent Planning Results}
\label{sec:experiments}

In the \textit{single-agent} setting, only one VLM-based agent acts as the food delivery courier across nine city maps.

\begin{table*}[t]
\centering
\caption{
Fine-grained evaluation of model capabilities across three dimensions:
High-level Planning, Resource Management, and Physical/Environmental Adaptation.
Arrows indicate whether higher ($\uparrow$) or lower ($\downarrow$) values are better.
}
\label{tab:fine_metrics}
\vspace{-5pt}

\footnotesize
\setlength{\tabcolsep}{4.5pt}     
\renewcommand{\arraystretch}{1.05}

\begin{tabular*}{\textwidth}{@{\extracolsep{\fill}} l|cccc|ccc|ccc}
\toprule
\multirow{2}{*}{\textbf{Model}} &
\multicolumn{4}{c|}{\textbf{Planning}} &
\multicolumn{3}{c|}{\textbf{Resources}} &
\multicolumn{3}{c}{\textbf{Physical \& Env.}} \\
\cmidrule(lr){2-5} \cmidrule(lr){6-8} \cmidrule(l){9-11}
& Order$\uparrow$ & OnTime$\uparrow$ & TimeEff$\uparrow$ & Active$\uparrow$ &
Stamina$\downarrow$ & Interrupts$\downarrow$ & Prevention$\uparrow$ &
Violations$\downarrow$ & Food$\uparrow$ & Cust$\uparrow$ \\
\midrule
GPT-5                    & 3.38 & 0.34 & 0.89 & 0.56 & 1.13 & 1.17 & 0.75 & 0.72 & 3.93 & 3.96 \\
GPT-4o                   & 3.36 & 0.38 & 0.54 & 0.58 & 1.28 & 1.61 & 0.66 & 0.69 & 3.82 & 3.94 \\
Claude-3.7-Sonnet        & \textbf{3.51} & \textbf{0.44} & 0.91 & \textbf{0.59} & 1.02 & \textbf{1.04} & \textbf{0.79} & \textbf{0.62} & 4.09 & \textbf{4.02} \\
Gemini-2.5-Flash         & 3.31 & 0.27 & \textbf{0.98} & 0.54 & 1.24 & 1.42 & 0.62 & 0.75 & 3.93 & 3.86 \\
\midrule
Qwen2.5-VL-72B-Ins       & 3.12 & 0.17 & 0.40 & 0.53 & 1.38 & 1.50 & 0.53 & 0.70 & \textbf{4.10} & 3.73 \\
Qwen2.5-VL-32B-Ins       & 3.43 & 0.16 & 0.48 & 0.47 & \textbf{0.98} & 1.05 & 0.74 & 0.65 & 3.87 & 3.48 \\
LLaMA-3.2-90B-Vision-Ins & 3.31 & 0.04 & 0.54 & 0.53 & 1.39 & 1.66 & 0.59 & 0.69 & 3.98 & 3.45 \\
\midrule
Human                    & 3.09 & 0.51 & 2.90 & 0.94 & 2.39 & 0.91 & 0.91 & 0.61 & 4.29 & 4.06 \\
\bottomrule
\end{tabular*}
\vspace{-10pt}
\end{table*}

\subsubsection{Global Performance}

Table~\ref{tab:global_performance} summarizes the net profits earned over a 2 virtual-hour episode across models and city sizes. \textbf{\textit{Closed-source models consistently achieve higher net profit than open-source models}}, with Claude-3.7-Sonnet achieving the highest net profit across all city sizes. Its relatively better performance in large cities reflects an advantage in handling long-horizon tasks, which involve longer delivery routes and more complex routing decisions. In contrast, many open-source models even incur losses in these cities. We also observe that closed-source models tend to have higher expenses, but much of this reflects \textbf{\textit{strategic investment}} for future deliveries (\eg, tool purchases), ultimately yielding higher profits. Nonetheless, \textbf{\textit{humans still outperform all models by a wide margin}} across all city sizes. On average, they earn over~\$50/hour, whereas the best model reaches only about~\$30/hour. We analyze this gap via a multi-dimensional breakdown.



\subsubsection{Fine-grained Analysis}
Table~\ref{tab:fine_metrics} presents the detailed results of the fine-grained trajectory-level analysis. Our key findings are as follows:


\begin{itemize}[leftmargin=10pt]

\item \textbf{Agents struggle to exploit temporal overlap compared with humans.} Agents fail to utilize their 2-hour window efficiently, often idling between actions (\eg, waiting to charge an e-scooter) instead of performing tasks concurrently (\eg, picking up food while charging), thereby wasting considerable time.
They tend to deliver orders sequentially rather than leveraging spatiotemporal alignment to complete multiple deliveries in parallel.
Consequently, their active-time and time efficiency remain substantially lower than those of humans.

\item \textbf{Agents remain less self-sustaining, often neglecting resource management and preventive actions.}
Most agents experience more than one interruption per hour due to stamina or battery depletion, and their proactive prevention ratios remain far below human results.
Even stronger models, such as Claude-3.7-Sonnet, often over-replenish when resources are sufficient and fail to act when depletion is imminent.

\item \textbf{Agents struggle to handle implicit, environment-dependent constraints.}
They often~overlook many implicit rules in delivery, choosing improper placement or transport methods that degrade food quality and trigger customer complaints (\eg, placing ice cream with hot food, causing it to melt). These constraint violations remain frequent, with both food and customer ratings staying relatively low, ultimately reducing their income.

\end{itemize}


\subsection{Multi-Agent Planning Results}

\noindent We further test VLM-based agents in multi-agent settings, where competition and collaboration naturally emerge.

\subsubsection{Global Performance}
We report model’s average net profit across all multi-agent group configurations on the \texttt{medium-20roads} map, as shown in Table~\ref{tab:multi_agent_profit}. Most models show a decline in profit when transitioning from the single-agent setting (without any competition or coordination) to multi-agent conditions.
Notably, GPT-4o exhibits the steepest drop. Compared to the purely competitive setup, all models except GPT-5 benefit from small-team cooperation, though their performance still remains well below the single-agent case.



\begin{table}[htbp]
\centering
\caption{
Multi-agent evaluation of average hourly net profit ($\bar{P}$) under five regimes: single-agent (1×1), fully competitive (8×1), and three cooperative structures (4×2, 2×4, 1×8). Underlines indicate the best-performing multi-agent configuration for each model.
}
\label{tab:multi_agent_profit}
\vspace{-5pt}
\scalebox{0.78}{
\renewcommand{\arraystretch}{1.05}
\begin{tabular}{l|c|cccc}
\toprule
\multirow{2}{*}{\textbf{Model}} &
\multicolumn{5}{c}{\textbf{Per-Agent Hourly Net Profit} ($\bar{P}$, \$/h)} \\
\cmidrule(l){2-6}
& (1$\times$1) &
\textbf{8$\times$1} &
\textbf{4$\times$2} &
\textbf{2$\times$4} &
\textbf{1$\times$8} \\
\midrule
GPT-5                    & \$27.3 & \underline{\$20.5} & \$19.5 & \$8.7 & \$16.5 \\
GPT-4o                   & \$16.9 & \$5.3 & \$5.5 & \$5.0 & \underline{\$6.9} \\
Claude-3.7-Sonnet        & \$31.7 & \$14.2 & \underline{\$22.6} & \$10.4 & \$9.6 \\
Gemini-2.5-Flash         & \$28.4 & \$21.2 & \underline{\$24.3} & \$12.6 & \$15.1 \\
\midrule
Qwen2.5-VL-72B-Ins       & \$10.1 & \$4.5 & \$7.0 & \underline{\$8.7} & \$5.8 \\
Qwen2.5-VL-32B-Ins       & \$6.0  & \$3.0 & \underline{\$4.6} & \$3.4 & \$1.4 \\
LLaMA-3.2-90B-Vision-Ins & \$1.4  & \$1.4 & \underline{\$2.0} & \$1.3 & \$1.5 \\
\bottomrule
\end{tabular}}
\vspace{-10pt}
\end{table}

\subsubsection{Impact of Team Size}
We analyze how team sizes affect coordination and interaction. As shown in Table~\ref{tab:multi_agent_profit}, \textbf{\textit{most models perform best in pairs, but some show declines as team size grows}}, especially in the four-agent setting. Although interaction events (\eg messaging or help requests) rise with team size, they also increase coordination overhead, as agents must manage more potential help requests alongside their own tasks, making it harder to prioritize effectively (\eg, accepting help requests but forgetting to act). The detailed change in interaction frequency is provided in the Appendix~\ref{appendix:interactionfrequency}.




\subsection{Agent Planning-Style Analysis}
\begin{figure}[ht]
    \centering
    \includegraphics[width=0.94\linewidth]{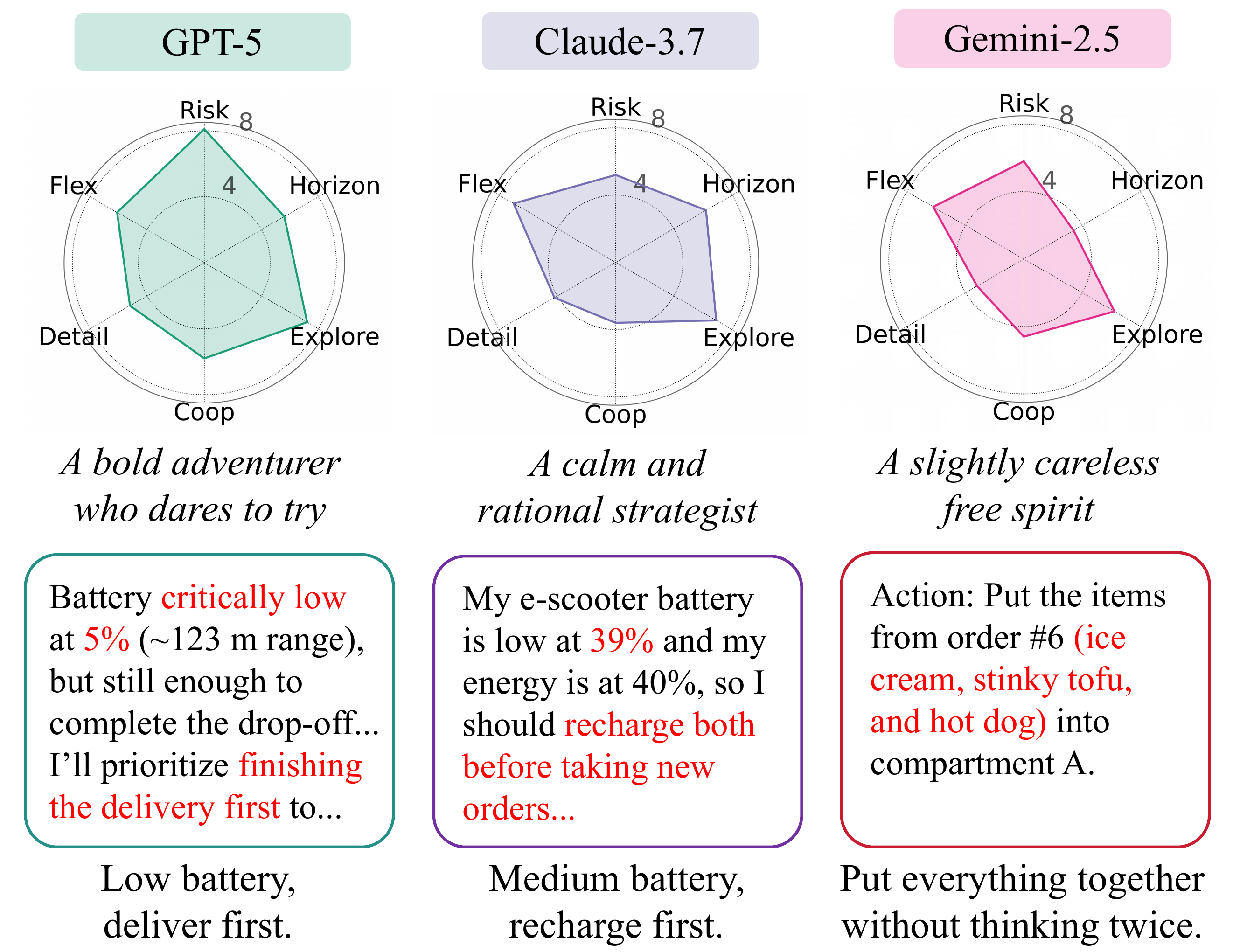}
    \vspace{-5pt}
    \caption{Comparison of model planning styles across six behavior dimensions, with example outputs provided as case studies.}
    \label{fig:planning_style}
    \vspace{-10pt}
\end{figure}

During both single- and multi-agent evaluations, we observe distinct decision-making and planning styles across models. For instance, Claude behaves more cautiously, choosing to head to a charging station once the e-scooter battery is low and pausing other tasks, whereas GPT-5 is more aggressive, often completing deliveries even with a nearly depleted battery. To further analyze model behavior in constraint-dense, real-world-like environments, we randomly sample delivery trajectories from each model and pair them with their outcomes. GPT-4o then evaluates each decision step across six dimensions on a 0–10 scale, including
\textit{Risk} (how aggressive the decision is),
\textit{Horizon} (preference for long-term planning or short-term gains),
\textit{Explore} (tendency to try new strategies),
\textit{Coop} (willingness to cooperate with others),
\textit{Detail} (attention to operational and contextual factors), and
\textit{Flex} (frequency of plan adjustments). Dimensions irrelevant to a given step are skipped. Figure~\ref{fig:planning_style} presents representative models with their planning styles and example outputs, and the full set of model evaluations, including action patterns, transportation modes, and spending distributions, is provided in the Appendix~\ref{appendix:modelbehavior}.


\vspace{-5pt}

\subsection{Context Engineering and Fine-tuning Effects}
\label{subsec:context_sft}

We evaluate two widely-used strategies for improving performance: Context Engineering and Supervised Fine-tuning (SFT) with human demonstrations, along with a baseline where the model outputs only raw actions without explicit planning for reference.
All evaluations in this section are conducted on the \texttt{medium-20roads} map.

\paragraph{Context Engineering.} Context Engineering aims to enhance model reasoning through self-reflection on prior experience and environmental feedback. We evaluate two methods: Agentic Context Engineering (ACE~\citep{zhang2025agentic}) and Dynamic Cheatsheet (DC~\citep{suzgun2025dynamic}). Each model undergoes a 4-hour warm-up phase, during which it updates an internal memory by summarizing key patterns from its past trajectories.
This memory is then frozen for evaluation. As shown in Table~4, context engineering consistently improves performance for GPT-5 and Claude-3.7-Sonnet, while the weaker open-source model Qwen2.5-VL-72B benefits little, with ACE even leading to a decline. Examples of the models' memory summaries are provided in the Appendix~\ref{appendix:contextstudy}.

\paragraph{Supervised Fine-tuning.} We fine-tune the open-source model LLaVA-OneVision-8B~\citep{an2025llava} on 9 human delivery trajectories (2,110 observation–action pairs) collected from the best-performing human on each map. We compare three variants: (i) the original pretrained model, (ii) a model fine-tuned directly on human actions, and (iii) a model fine-tuned on \textit{annotated} human actions, where each action is enriched with \textit{reasoning}, \textit{reflection}, and \textit{future plans} generated by GPT-4o. All variants are trained for 3 epochs. The model fine-tuned on raw human actions exhibits more human-like behaviors (e.g., bundling orders) but performs worse, often imitating patterns without understanding preconditions (\eg charging without reaching a station).
In contrast, the annotated variant performs better, achieving higher profits and learning human-like parallel task strategies that significantly improve time efficiency and active ratio. The fine-grained analysis can be found in Appendix~\ref{appendix:context_sft_finegrained}.

\begin{table}[t]
\centering
\caption{
Comparative results of context engineering and supervised fine-tuning. Green and red highlights improvements and regressions over the with-Plan baseline, respectively.
}
\vspace{-5pt}
\label{tab:ablation-noplan}
\begingroup
\renewcommand{\arraystretch}{1.05}
\resizebox{0.9\columnwidth}{!}{
\begin{tabular}{l|ccc}
\toprule
\textbf{Model} & $\bar{P}$ & $E$ & $C$ \\
\midrule
GPT-5 (with Plan)                   & \$27.3 & \$38.8 & \$11.5 \\
GPT-5 (w/o Plan)                    & \cellcolor{red!12}\$8.6  & \cellcolor{red!12}\$16.8 & \cellcolor{green!12}\$8.2 \\
GPT-5 (with Plan + ACE)             & \cellcolor{green!12}\$33.2 & \cellcolor{green!12}\$46.1 & \cellcolor{red!12}\$12.9 \\
GPT-5 (with Plan + DC)              & \cellcolor{green!12}\$36.2 & \cellcolor{green!12}\$47.3 & \cellcolor{green!12}\$11.2 \\
\midrule
Claude-3.7-Sonnet (with Plan)       & \$31.7 & \$51.6 & \$19.9 \\
Claude-3.7-Sonnet (w/o Plan)        & \cellcolor{red!12}\$19.2 & \cellcolor{red!12}\$25.6 & \cellcolor{green!12}\$6.3 \\
Claude-3.7-Sonnet (with Plan + ACE) & \cellcolor{green!12}\$40.5 & \cellcolor{green!12}\$56.3 & \cellcolor{green!12}\$15.8 \\
Claude-3.7-Sonnet (with Plan + DC)  & \cellcolor{green!12}\$44.5 & \cellcolor{green!12}\$57.1 & \cellcolor{green!12}\$12.6 \\
\midrule
Qwen2.5-VL-72B (with Plan)          & \$2.3  & \$14.0 & \$11.7 \\
Qwen2.5-VL-72B (w/o Plan)           & \cellcolor{red!12}\$2.0   & \cellcolor{red!12}\$10.8 & \cellcolor{green!12}\$8.8 \\
Qwen2.5-VL-72B (with Plan + ACE)    & \cellcolor{red!12}\$0.1   & \cellcolor{green!12}\$14.3 & \cellcolor{red!12}\$14.2 \\
Qwen2.5-VL-72B (with Plan + DC)     & \cellcolor{green!12}\$3.2  & \cellcolor{green!12}\$16.6 & \cellcolor{red!12}\$13.4 \\
\midrule
LLaVA-OneVision-8B (original)        & -\$7.2 & \$4.4  & \$11.6 \\
LLaVA-OneVision-8B (raw-action-ft)   & \cellcolor{red!12}-\$7.8 & \cellcolor{green!12}\$7.2 & \cellcolor{red!12}\$15.0 \\
LLaVA-OneVision-8B (annotated-ft)    & \cellcolor{green!12}\$3.2 & \cellcolor{green!12}\$12.7 & \cellcolor{green!12}\$9.5 \\
\bottomrule
\end{tabular}}
\endgroup
\end{table}

\section{Conclusion}
We introduced \deliverybench, an embodied benchmark to evaluate VLM-based agents under realistic, long-horizon delivery scenarios. 
In the grounded food-delivery profession, agents must maximize long-term profit while simultaneously handling spatial, temporal, resource, physical, economic, and social constraints. 
By instantiating these demands in simulated 3D cities with diverse layouts, multiple transportation modes, and both single- and multi-agent regimes, \deliverybench provided a more faithful and diagnostic testbed for studying constraint-aware planning.
Our experiments across nine cities with a range of state-of-the-art VLMs reveal a substantial gap to human couriers, exhibiting their short-sighted behavior and frequent break of basic commonsense constraints. 
Besides, different models display distinct behavioral personalities, highlighting both diversity and brittleness in current VLM-based agents.


{
    \small
    \bibliographystyle{ieeenat_fullname}
    \bibliography{main}
}

\clearpage

\appendix
\section{Future Research Directions}
\label{appendix:future_direction}
\deliverybench simulates real-world food-delivery task, which naturally involves long-horizon objectives (\eg maximizing net profit) intertwined with diverse physical, social, and economic constraints, providing a testbed that more faithfully reflects the complexity of real-world decision-making. As a next step, we aim to further extend this platform in several important directions:

\paragraph{Real-time reasoning.}
In the current setup, the simulator pauses the environment whenever the model is ``thinking'': order timers, battery levels, food freshness, and other dynamic states are frozen. In contrast, real-world decision-making unfolds in a continuously evolving environment, where time keeps progressing and other entities (\eg couriers, pedestrians, customers) act in parallel. We plan to support real-time planning in future versions, where agents must reason within this dynamic setting and adapt to ongoing temporal and environmental changes (e.g., adjusting their trajectory in real time to avoid pedestrians).

\paragraph{Learning from interaction data.} Although \deliverybench currently serves primarily as an evaluation benchmark, the platform naturally supports collecting rich interaction data at scale. Such data can be used to study how different learning paradigms, including reinforcement learning, imitation learning, and memory-augmented agents, adapt to our long-horizon delivery task. As shown in Section~\ref{subsec:context_sft}, we conduct preliminary experiments using basic context engineering and small-scale supervised fine-tuning from human demonstrations, but there remains substantial room for further investigation, especially in understanding how these methods scale as data and model size increase.
\section{\deliverybench Details}
\label{appendix:deliverybench}

We provide additional details of \deliverybench, including map construction, transportation and POI design, and several task-specific mechanisms (\eg food categories).

\subsection{City Maps and Spatial Layout}
\begin{figure}[ht]
    \centering
    \includegraphics[width=0.94\linewidth]{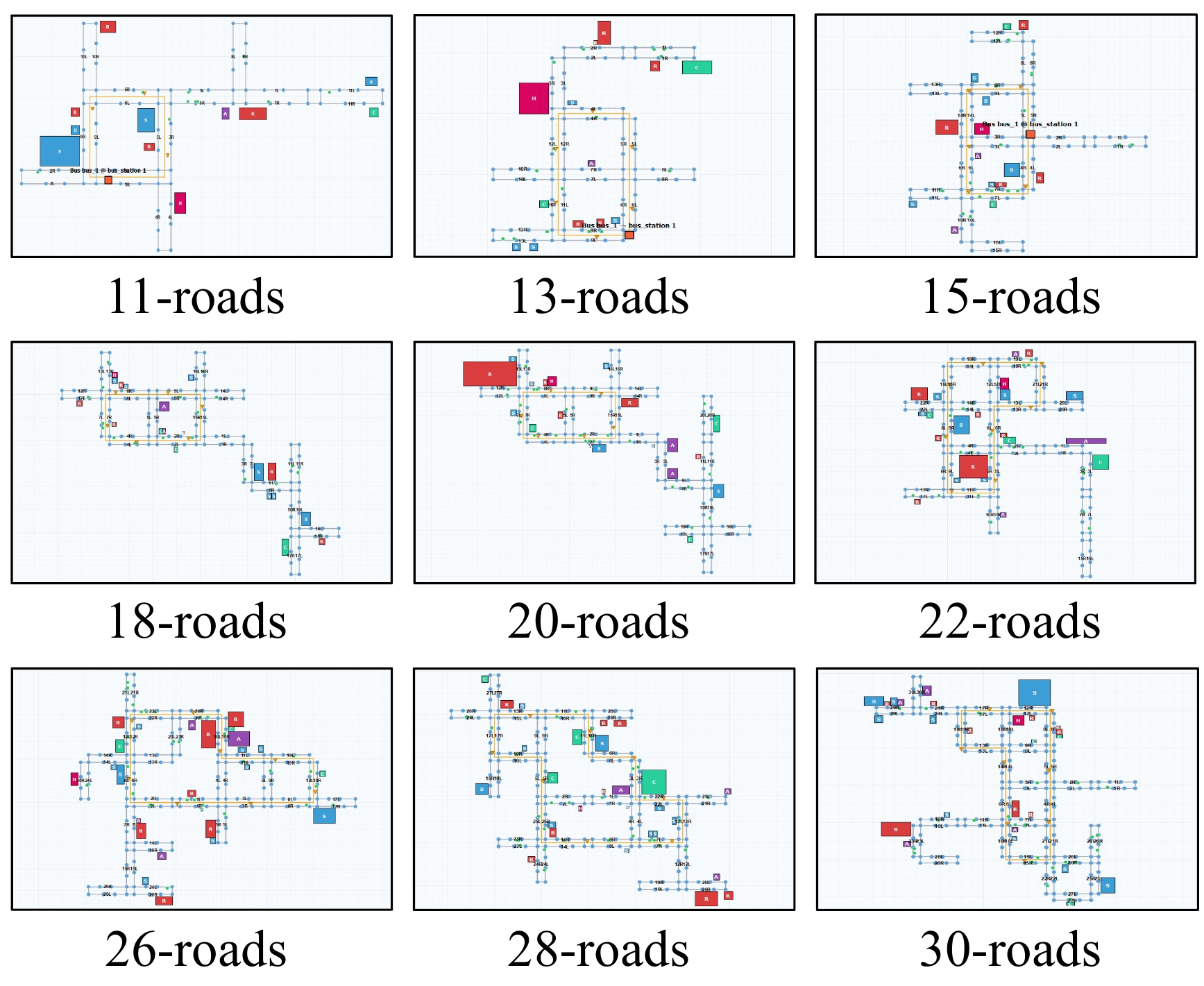}
    \vspace{-5pt}
    \caption{Overview of the nine procedurally constructed city maps used in our experiments.}
    \label{fig:map_layouts}
    \vspace{-10pt}
\end{figure}

We construct nine city maps spanning three difficulty levels: \emph{small} (11–15 roads), \emph{medium} (16–25 roads), and \emph{large} (26–30 roads), with three maps in each category. Every map contains a diverse set of POIs distributed across the road network, sampled under a uniform spatial density such that larger maps naturally include more POIs. For each city, we select the largest inscribed loop as the bus route, evenly place bus stops along it, and deploy a single bus that continuously travels on this route. The overall spatial layouts of the maps are illustrated in Figure~\ref{fig:map_layouts}, and the POI statistics for each map are summarized in Table~\ref{tab:poi_stats}.

\subsection{Transportation Modes}
We provide multiple transportation modes, including e-scooter, walking, driving, and public transit such as buses. These modes differ in speed, stamina consumption, and additional costs (e.g., bus fares, car rental fees), requiring the model to make context-dependent trade-offs.
A summary of these transportation modes is provided in Table~\ref{tab:transport_modes}.

\vspace{-5pt}
\begin{table}[htbp]
\centering
\caption{Different transportation modes in DeliveryBench.}
\label{tab:transport_modes}
\vspace{-5pt}
\resizebox{\linewidth}{!}{%
\renewcommand{\arraystretch}{1.05}
\begin{tabular}{lccc}
\toprule
\textbf{Mode} & \textbf{Speed (m/s)} & \textbf{Stamina (\%/m)} & \textbf{Extra Cost} \\
\midrule
walk          & 2.0   & 0.08  & -- \\
e-scooter     & 6.0   & 0.01  & battery $0.04\%\!/$m \\
drag e-scooter & 1.5   & 0.10  & -- \\
car           & 12.0  & 0.008 & rental \$1.0/min \\
bus           & 10.0  & 0.006 & \$1 fare \\
\bottomrule
\end{tabular}%
}
\vspace{-6pt}
\end{table}

\begin{table*}[t]
\centering
\caption{Counts of points of interest (POIs) on each \deliverybench map.}
\vspace{-5px}
\label{tab:poi_stats}
\resizebox{0.9\linewidth}{!}{%
\begin{tabular}{lccccccccc}
\toprule
Size & \#Roads & Restaurant & Store & Rest Area & Car Rental & Hospital & Charging Station & Bus Station & Bus Route \\
\midrule
\multirow{3}{*}{small}  & 11 & 4 & 4 & 1 & 1 & 1 & 10 & 4 & 1 \\
                        & 13 & 5 & 4 & 1 & 2 & 1 & 15 & 6 & 1 \\
                        & 15 & 4 & 5 & 2 & 2 & 1 & 18 & 6 & 1 \\
\midrule
\multirow{3}{*}{medium} & 18 & 6 & 7 & 2 & 3 & 1 & 20 & 6 & 1 \\
                        & 20 & 5 & 7 & 3 & 3 & 1 & 24 & 6 & 1 \\
                        & 22 & 7 & 7 & 3 & 3 & 1 & 22 & 8 & 1 \\
\midrule
\multirow{3}{*}{large}  & 26 & 7 & 9 & 4 & 4 & 1 & 29 & 8 & 1 \\
                        & 28 & 8 & 11 & 3 & 4 & 1 & 29 & 8 & 1 \\
                        & 30 & 9 & 9 & 4 & 3 & 1 & 24 & 8 & 1 \\
\bottomrule
\end{tabular}%
}
\end{table*}

\subsection{Points of Interest}
Our constructed city includes various POIs, each serving distinct functions.  
Agents must navigate the city and interact with these POIs to accomplish different subtasks.

\paragraph{Restaurant.}
Restaurants serve as the pickup locations for delivery orders. Once an order is accepted, the restaurant begins food preparation. When the meal is ready, its state (e.g., temperature or freshness) starts changing over time, and the agent can visit the restaurant to collect it.

\paragraph{Store.}
Stores provide agents with access to purchasable items, including energy drinks, e-scooter batteries, and food-preservation tools such as ice packs and heat packs. The prices and functions of these items are listed in Table~\ref{tab:store_items}.

\begin{table}[htbp]
\centering
\caption{Prices and functions of store items.}
\label{tab:store_items}
\vspace{-5pt}
\resizebox{\linewidth}{!}{%
\renewcommand{\arraystretch}{1.05}
\begin{tabular}{lcc}
\toprule
\textbf{Item} & \textbf{Price (\$)} & \textbf{Function} \\
\midrule
Energy Drink     & 6   & Restore 50\% of stamina \\
E-Scooter Battery  & 10  & Fully recharge e-scooter battery \\
Ice Pack         & 3   & Cool food temperature \\
Heat Pack        & 3   & Heat food temperature \\
\bottomrule
\end{tabular}%
}
\vspace{-6pt}
\end{table}

\paragraph{Rest Area.}
Rest areas provide couriers with a place to recover stamina, allowing agents to restore 10\% of their stamina per minute at no cost while resting.

\paragraph{Car Rental.}
Car rental stations allow agents to rent and return cars. An agent can pick up a car at any rental station and return it to any other. Rental fees are time-based and cost \$0.5 per minute, even when the vehicle is not in use.

\paragraph{Hospital.}
Hospitals handle agent recovery when stamina is fully depleted. An agent who collapses is automatically sent to a hospital for a 30-minute recovery process, during which no actions can be performed and a \$5 service fee is charged. All environment dynamics, such as order timers and food freshness, continue to progress normally. After recovery, the agent resumes work starting from the hospital.

\paragraph{Charging Station.}
Charging stations provide recharging services for agents’ e-scooters, with each station able to serve only one scooter at a time. The charging cost is \$0.05 per unit of battery, and the charging speed is 10 units per minute. Agents may stop charging and retrieve their e-scooters at any time.

\paragraph{Bus Station.}
Bus stations allow agents to wait for the arriving bus and board it when it reaches the stop. Upon arrival, agents may pay a \$1 ticket fee and ride the bus to any other station on the route.

\begin{figure*}[ht]
    \centering
    \includegraphics[width=0.9\linewidth]{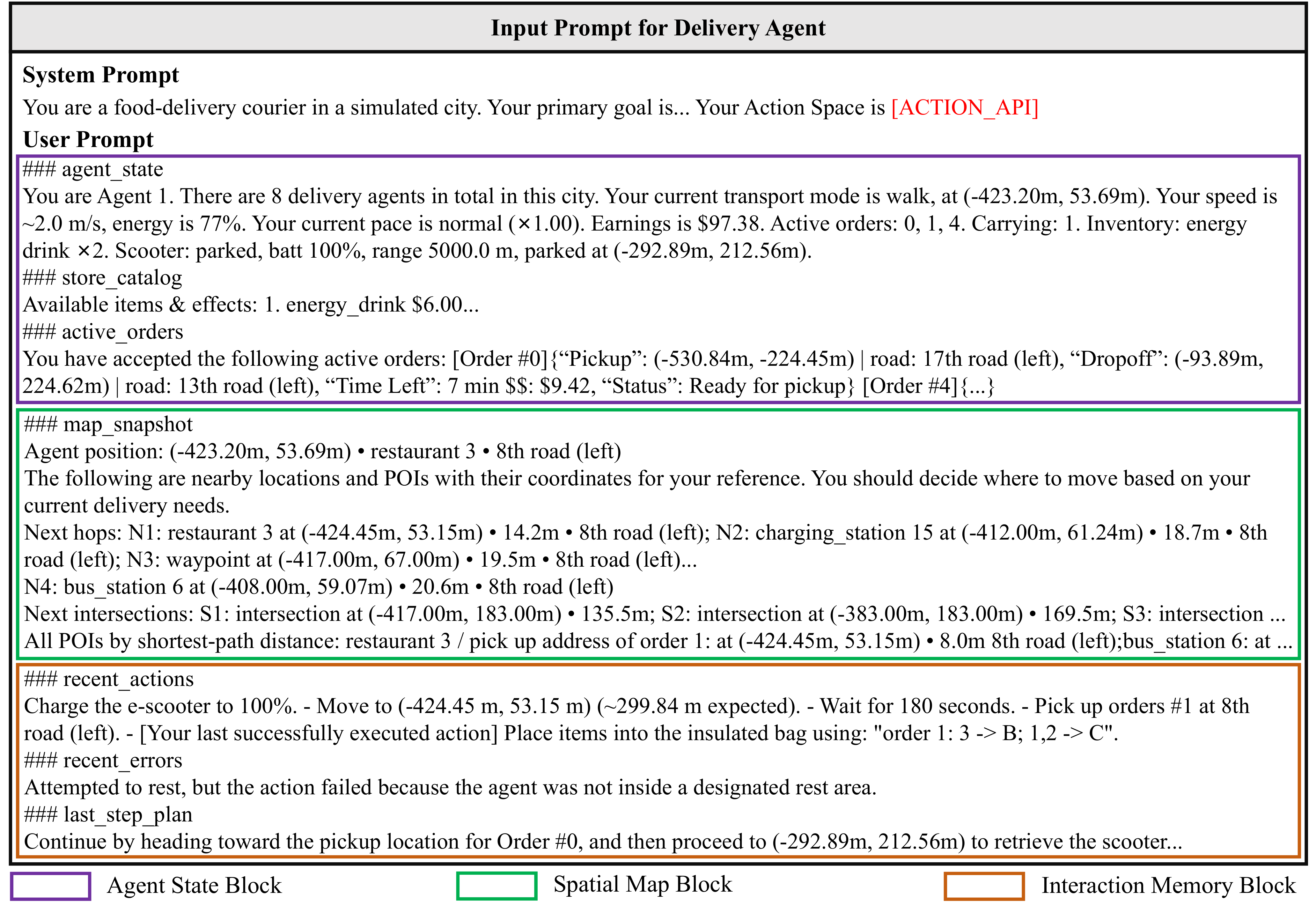}
    \vspace{-5pt}
    \caption{Overview of the input prompt used by delivery agents}
    \label{fig:input_prompt_delivery_agent}
    \vspace{-10pt}
\end{figure*}

\subsection{Food Attributes}
We simulate 22 food types, each with a preparation time and several quality-related attributes.
These attributes influence how the food evolves during delivery and influence the agent’s strategy. 
The main factors include temperature dynamics, fragility, and odor sensitivity.

\paragraph{Temperature Dynamics.}
Temperature is the most influential factor affecting food quality. After preparation, a food item’s temperature evolves according to a lightweight thermodynamic model that simulates heat exchange with its surroundings. Each item has a temperature $T_i$ and heat capacity $C_i$, while each storage compartment has an air node with temperature $T_a$ and a small heat capacity $C_{ab}$. Items outside the insulated bag exchange heat with ambient air, whereas items inside the bag primarily exchange heat with others in the same compartment. We update temperatures using a discrete heat-exchange rule with timestep $\Delta t$:
\begin{align}
S &= \sum_i C_i (T_i - T_a), \\
T_a^{\text{new}} &= T_a + \alpha \frac{S}{C_{ab}}, \\
T_i^{\text{new}} &= T_i + \alpha (T_a - T_i),
\end{align}
where $S$ denotes the net heat flow from the food items to the air node. The coefficient $\alpha = \Delta t / \tau_{\mathrm{ex}}$ controls the exchange rate and is clipped to $\alpha \le 0.5$ for numerical stability, while $\tau_{\mathrm{ex}}$ determines the effective speed of heat transfer.

\paragraph{Fragility.}
Items such as cakes and soups are sensitive to movement and require gentle handling.
Actions involving rapid movement (e.g., riding an e-scooter at high speed or running) introduce a risk of damaging these items. 
Each fragile item accumulates a fragility score when subjected to excessive vibration or acceleration. 
Once the accumulated damage exceeds a threshold, the food is considered ruined.

\paragraph{Odor Sensitivity.}
Strong-smelling foods (\eg stinky tofu or durian) can affect other items stored in close proximity. When such foods are placed in the same insulated compartment as milder items, prolonged storage can lead to odor transfer. We model this using a simple odor-mixing mechanism. Each food item maintains an odor level $o_i \in [0,1]$, and items within the same compartment gradually converge toward the highest odor level present in that compartment:
\[
o_i^{\text{new}} = o_i + \alpha \bigl(o_{\max} - o_i\bigr),
\]
where $o_{\max}$ is the maximum odor level among items in the compartment, and $\alpha$ is a small timestep-based update coefficient. If $o_{\max}=0$, no odor transfers.

\subsection{Order Attributes}
Orders serve as the fundamental task units in our simulation. Each order specifies a designated pickup restaurant, a drop-off address, a delivery time window, and an associated wage. Some orders may also include special customer requests, which agents must carefully consider during fulfillment. Upon successful delivery, the system automatically settles the base wage and applies any additional bonuses based on customer ratings.

\paragraph{Delivery Methods.} Agents may choose from four delivery methods: leaving the item at the doorstep, calling the customer, knocking on the door, or handing the order directly to the customer. For face-to-face delivery, the agent must first locate the customer's actual position (e.g., “under the tree near the entrance”) and approach them to trigger the handoff. The other methods only require reaching the designated building entrance.
If the order includes no customer notes, any of the four methods is acceptable. However, if specific delivery instructions are provided, the agent must infer the most appropriate method from the context. For example, a note saying ``I’m in a meeting'' suggests the agent should leave the item at the door to avoid interruption, while high-value items may warrant direct handoff. Choosing an inappropriate delivery method can result in customer dissatisfaction and lower ratings.

\paragraph{Base Delivery Pay.}
Each delivery order includes a fixed base wage, which is granted in full if the agent completes the delivery within the specified time window or a short grace period (\eg 1 minute). For late deliveries, the base pay is proportionally reduced based on the delay duration, but never falls below 30\% of the original amount.

\paragraph{Customer Rating Bonus.}
Upon successful delivery, the customer provides a rating from 0 to 5 based on overall satisfaction. This rating influences the agent’s compensation through a bonus or penalty mechanism. The score reflects three main factors: total customer waiting time, food condition upon arrival, and the suitability of the chosen delivery method.
If the rating exceeds 3 stars, the agent receives a bonus of up to \$3. If the rating falls below 3 stars, a fixed \$2 penalty is applied.

\section{Agent Input–Output Specification}
\label{appendix:deliveryagent}
In this section, we specify the delivery agent’s input and output formats, along with its action space.

\begin{figure*}[ht]
    \centering
    \includegraphics[width=0.8\linewidth]{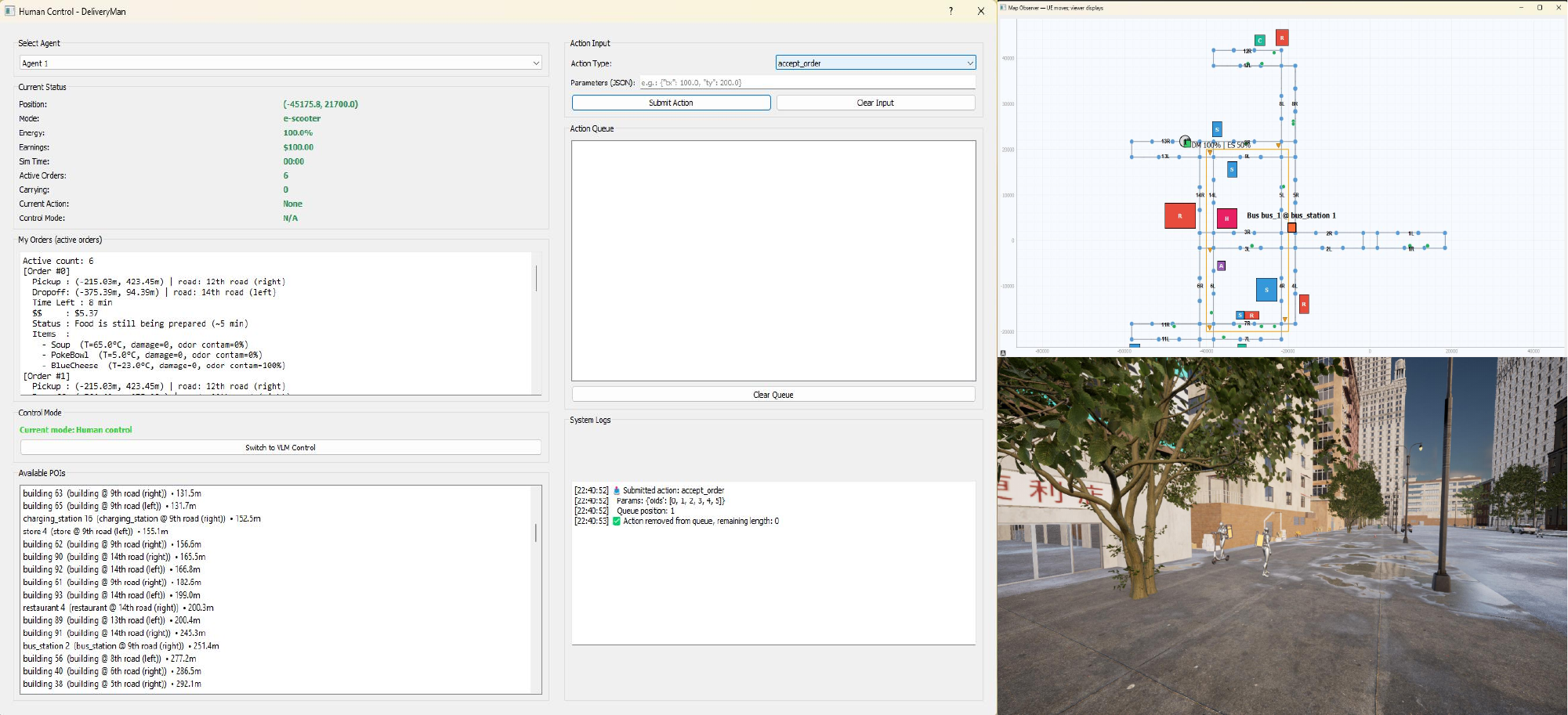}
    \vspace{-5pt}
    \caption{Human interaction GUI.}
    \label{fig:humanGUI}
    \vspace{-10pt}
\end{figure*}

\subsection{Input Prompt Structure}
At each decision step, the agent receives an input prompt that summarizes all information needed for planning and acting. 
The prompt consists of two parts: a \emph{System Prompt}, which remains fixed throughout the episode, and a \emph{User Prompt}, which is dynamically updated at every step. 
The System Prompt specifies the agent’s role in the simulated city, its primary delivery objective, and the available action space.
The User Prompt then provides three additional components: 
(i) an \emph{Agent State} block describing the agent’s current status, such as its location, transport mode, speed, energy level, and active orders; 
(ii) a \emph{Spatial Map} block encoding a compact map snapshot, including the next reachable waypoints, nearby intersections, and the locations of relevant POIs; 
and (iii) an \emph{Interaction Memory} block recording recent actions, the previous step’s plan, and any error messages from failed actions.
Sometimes the User Prompt also includes context-specific information; for example, arriving at a restaurant reveals the list of available pickups, and invoking an order-viewing action inserts the current order pool into the prompt. An example of the full prompt structure is shown in Figure~\ref{fig:input_prompt_delivery_agent}.

\subsection{Output Format}
\label{appendix:outputformat}

The agent follows a fixed structured format when producing its textual output. It first reflects on its recent memory and current state to formulate a \emph{Reflection and Reasoning} paragraph that explicitly articulates the thought process behind the current decision. Based on this reasoning, the agent then outputs an \emph{Action} specifying the concrete operation to execute. Finally, it provides a \emph{Future Plan} describing how it intends to proceed after completing the current action.

\subsection{Action Space}
\label{appendix:actionspace}

In DeliveryBench, the agent selects from a discrete action space of 30 actions, organized into several functional categories: 
(i) \textit{Movement actions} allow the agent to navigate across the city, either through high-level navigation commands that invoke the built-in shortest-path planner or through simple low-level motion steps (\eg stepping forward or turning around). 
(ii) \textit{Order-handling actions} support core delivery operations such as browsing the order pool, accepting orders, and completing drop-offs. 
(iii) \textit{Inventory and resource management actions} involve managing the agent’s internal resources, enabling it to regulate stamina, battery levels, and food conditions (\eg resting, inspecting the bag, consuming energy drinks or battery packs).
(iv) \textit{Social and collaboration actions} facilitate multi-agent assistance, including viewing or posting help requests, accepting cooperative tasks, and simple communication. 
(v) \textit{Transportation actions} allow the agent to switch transportation modes, rent or return vehicles, or use the public bus system.

\section{Human Data Collection}
\label{appendix:humandata}

To obtain a reasonable human performance reference and collect data for supervised fine-tuning, we recruited three human participants, each completing a two-hour delivery session independently. All experimental settings and evaluation protocols were kept identical to those used for the VLM agent. The resulting human trajectories were then augmented using GPT-4o to generate the corresponding reflection, reasoning, and future-plan annotations.

\begin{table*}[!h]
\centering
\caption{
Fine-grained metrics for delivery agents; arrows indicate whether higher ($\uparrow$) or lower ($\downarrow$) values are better.
}
\label{tab:metric_definitions}

{
\setlength{\aboverulesep}{0pt}   
\setlength{\belowrulesep}{0pt}   
\renewcommand{\arraystretch}{1.6}
\setlength{\tabcolsep}{6pt}
\small

\resizebox{0.85\textwidth}{!}{%
\begin{tabular}{@{}l|%
    >{\raggedright\arraybackslash}m{0.13\linewidth}%
    >{\raggedright\arraybackslash}m{0.60\linewidth}%
    c}
\toprule
\textbf{Dimension} & \textbf{Metric} & \textbf{Definition} & \textbf{Range} \\
\midrule

\textbf{Planning} & Order (Quality) $\uparrow$
& Average relative quality of the orders selected by the agent, evaluated based on delivery-deadline feasibility relative to distance, reward relative to cost, and the alignment between the order’s delivery route and the agent’s current trajectory. Candidate orders are scored and ranked within the pool, with higher-ranked orders indicating higher quality.
& [0, 5] \\ \cline{2-4}

& OnTime (Rate) $\uparrow$
& Proportion of selected orders delivered before their deadlines.
& [0, 1] \\ \cline{2-4}

& TimeEff (Time Efficiency) $\uparrow$
& Sum of effective delivery durations for all delivered orders, including periods where multiple orders are handled in parallel, divided by the total episode time. Values greater than 1 indicate that the agent frequently handles multiple orders in parallel, values close to 1 indicate that the agent is almost continuously engaged in deliveries, and values below 1 indicate substantial idle time between deliveries.
& [0, 1] \\ \cline{2-4}

& Active (Rate) $\uparrow$
& Fraction of time spent performing purposeful actions (\eg moving, picking up, delivering, recharging), excluding waiting or incapacitated periods.
& [0, 1] \\
\midrule

\textbf{Resources} & StaminaUse $\downarrow$
& Average stamina consumption per hour.
& $\ge 0$ \\ \cline{2-4}

& Interrupts $\downarrow$
& Number of forced interruptions per hour caused by resource depletion (\eg stamina or battery exhaustion).
& $\ge 0$\\ \cline{2-4}

& Prevention $\uparrow$
& Fraction of times the agent replenishes critical resources before they are depleted.
& [0, 1] \\
\midrule

\textbf{Physical \& Env.} & Violations $\downarrow$
& Proportion of orders that incur constraint violations, such as food-quality failures (\eg melting, breakage, or odor transfer).
& [0, 1] \\ \cline{2-4}

& FoodRate $\uparrow$
& Average rating of the food’s final quality upon delivery.
& [0, 5] \\ \cline{2-4}

& CustRate $\uparrow$
& Average customer rating for each delivered order, reflecting overall satisfaction with factors such as waiting time, delivery behavior, and food condition.
& [0, 5] \\
\bottomrule
\end{tabular}
} 
} 
\end{table*}

\subsection{Human Interaction GUI}
\label{appendix:humangui}

Human participants interacted with the environment via a custom-designed GUI that provides first-person observations, a map view, and contextual task information. Participants issued their actions directly through the interface. During delivery, the GUI displays real-time information such as the participant's remaining stamina, current location, and accumulated earnings. All human trajectories are automatically logged by the system. A detailed illustration of the GUI is provided in Figure~\ref{fig:humanGUI}.

\subsection{LLM-enhanced Annotation}
Since the human trajectories only record the actions chosen at each step, we use GPT-4o to reconstruct the full chain-of-thought annotations in the same structured format described in Appendix~\ref{appendix:outputformat}, ensuring consistency with the VLM agent’s outputs. 
For each human decision step, we provide GPT-4o with the corresponding observation and executed action, prompting the model to infer the underlying rationale behind the decision. 
We further supply the subsequent five human actions to GPT-4o, enabling it to generate the future plan aligned with those actions.

\section{Evaluation Details}
\label{appendix:evaluation}

\subsection{Fine-grained Metric Definitions}
\label{appendix:metrics}

To analyze agent behavior beyond final delivery profit, we adopt a set of fine-grained metrics that capture different aspects of long-horizon delivery performance. These metrics assess high-level planning (order selection, deadline handling, time utilization), resource management (stamina usage and proactive replenishment), and physical or environmental adaptation (food quality, constraint violations, customer satisfaction). Their formal definitions and computation methods are summarized in Table~\ref{tab:metric_definitions}.

\subsection{Planning Style Evaluation Prompts}
\label{appendix:llm_eval}

\begin{figure*}[ht]
    \centering
    \includegraphics[width=0.9\linewidth]{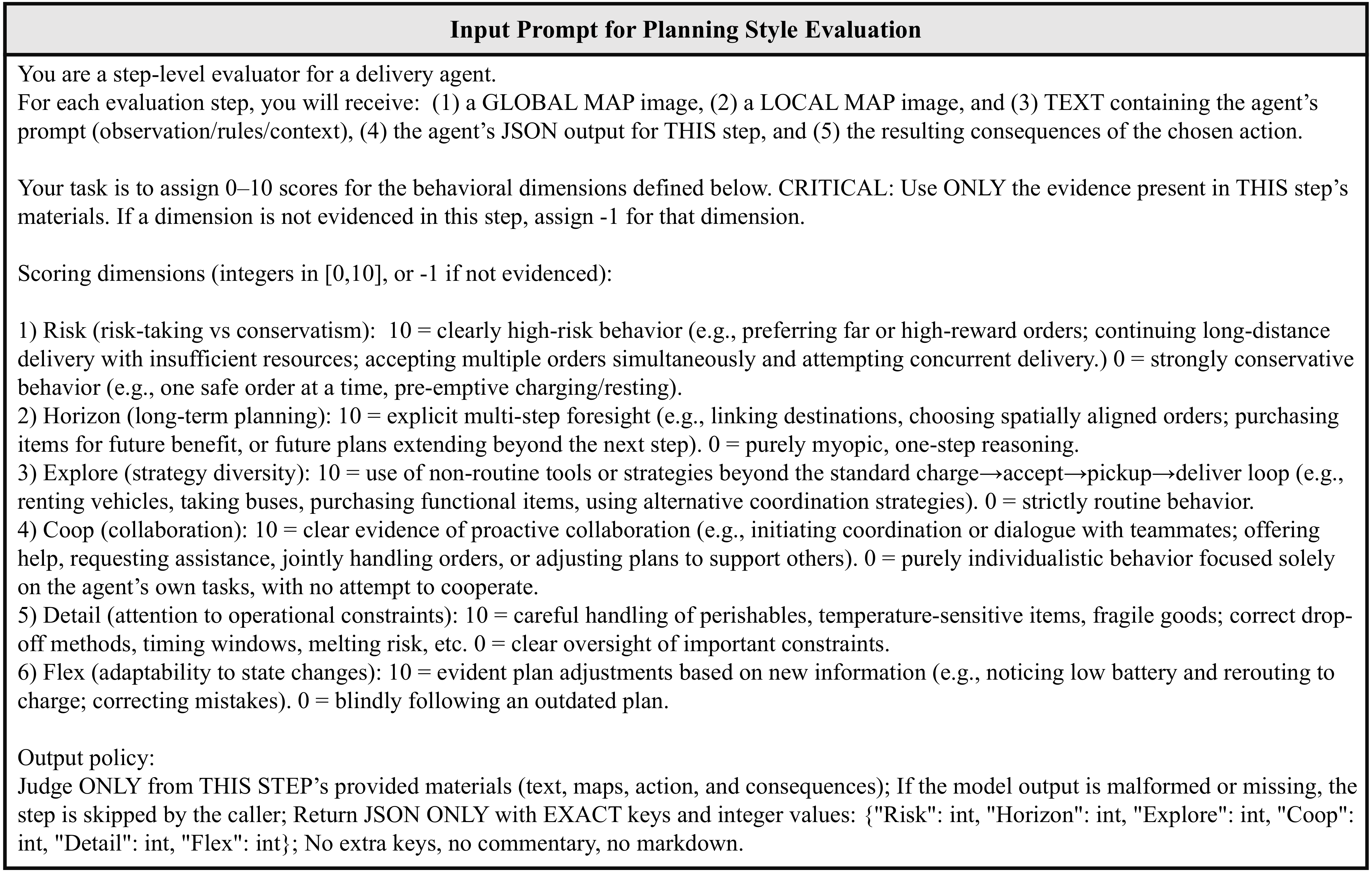}
    \vspace{-5pt}
    \caption{Prompt for planning style evaluation.}
    \label{fig:planning_style_prompt}
    \vspace{-10pt}
\end{figure*}

We use GPT-4o as an evaluator to assess the planning style exhibited by each model. At each evaluation step, GPT-4o is given (i) the current environment observation and (ii) the model’s full output, which includes the chosen action, its chain-of-thought rationale, and the resulting consequences of that decision (e.g., whether an accepted order later times out or whether the action leads to future battery depletion). GPT-4o then scores this decision across multiple planning dimensions. The complete evaluation prompt used for scoring is shown in Figure~\ref{fig:planning_style_prompt}. 
\section{Additional Experimental Results}
\label{appendix:experiments}

\subsection{Interaction Frequency with Team Size}
\label{appendix:interactionfrequency}

\begin{figure}[!h]
    \centering
    \includegraphics[width=0.6\linewidth]{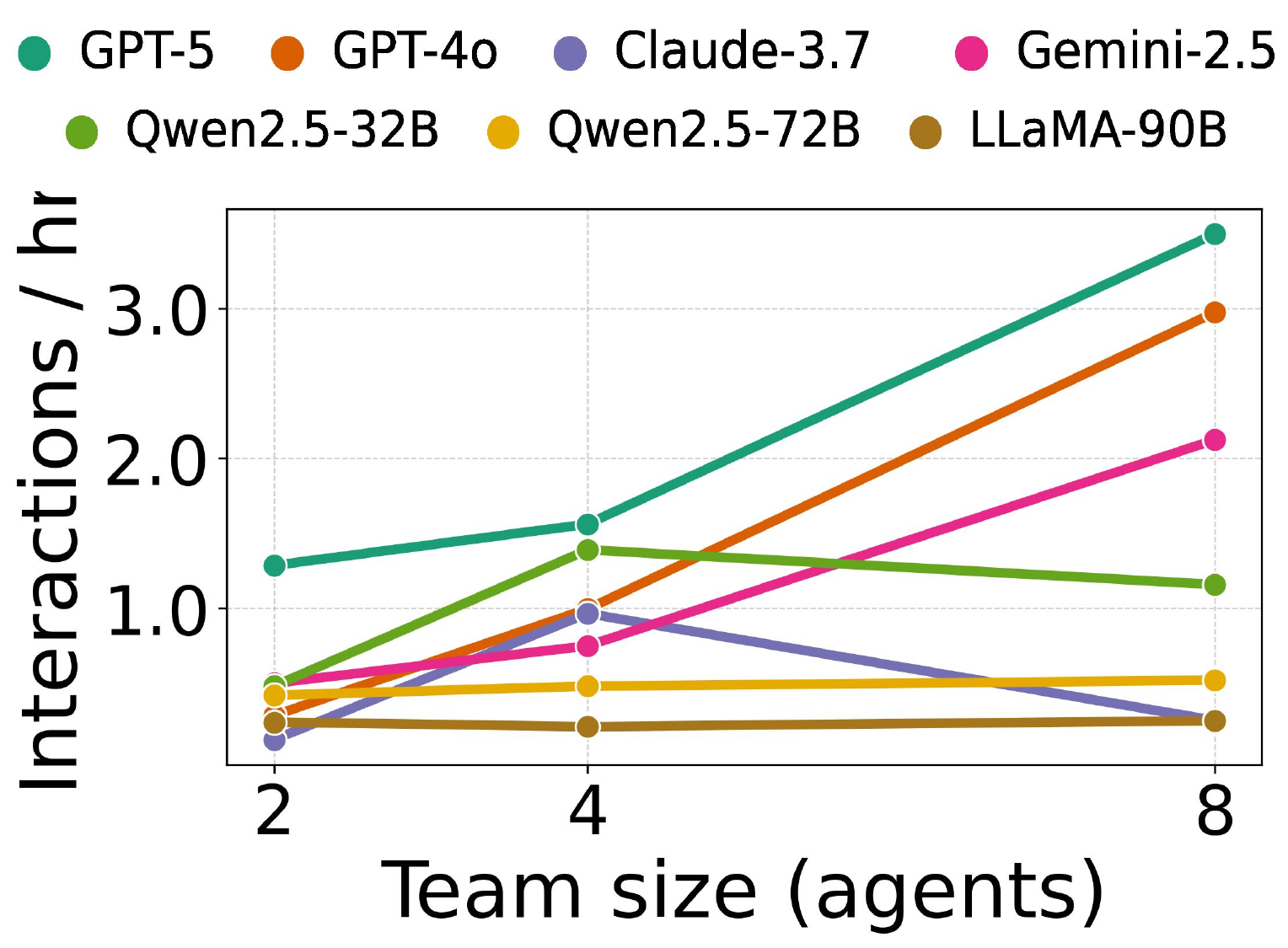}
    \caption{Interaction frequency across team sizes.}
    \label{fig:multi_interaction}
    \vspace{-10pt}
\end{figure}

In the multi-agent setting, we evaluate how interaction frequency among models changes with team size, as shown in Figure~\ref{fig:multi_interaction}. Although the communication rate tends to increase in larger teams, agents still interact only occasionally. However, this increase in communication does not improve task performance. As team size grows, coordination becomes more complex. Agents must balance maximizing their own utility with supporting their teammates, which makes effective cooperation more difficult. As a result, agents often overreact to teammate requests and abandon their own tasks, or they promise help but fail to follow 
through, leaving both sides stalled.

\subsection{Model Behaviors and Planning Styles}
\label{appendix:modelbehavior}

\begin{figure}[ht]
    \centering
    \includegraphics[width=\linewidth]{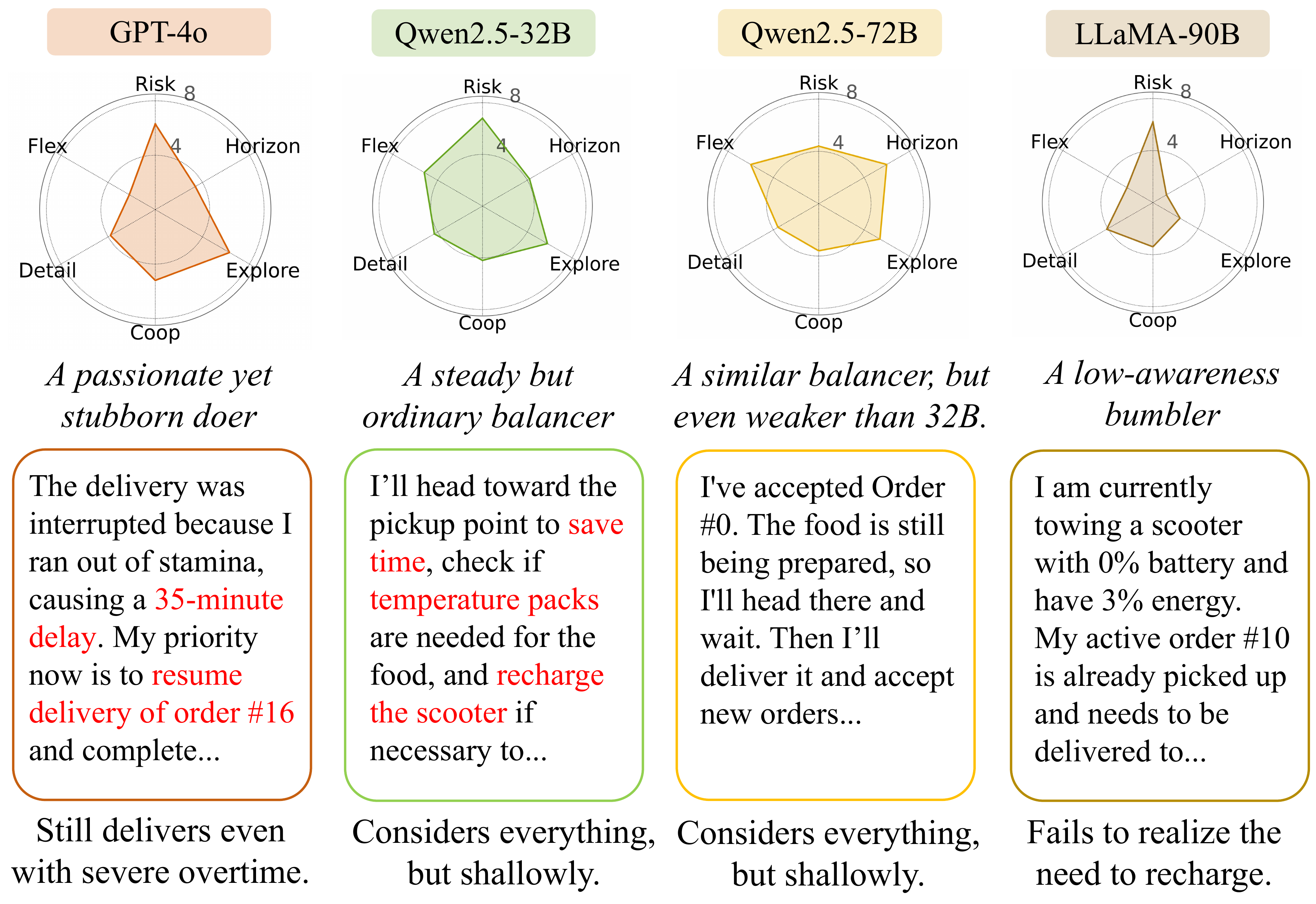}
    \vspace{-10pt}
    \caption{Planning style visualizations for the remaining four models, complementing the examples shown in Figure~\ref{fig:planning_style}.}
    \label{fig:planning_style_4}
    \vspace{-10pt}
\end{figure}

\begin{figure*}[!h]
    \centering
    \includegraphics[width=\linewidth]{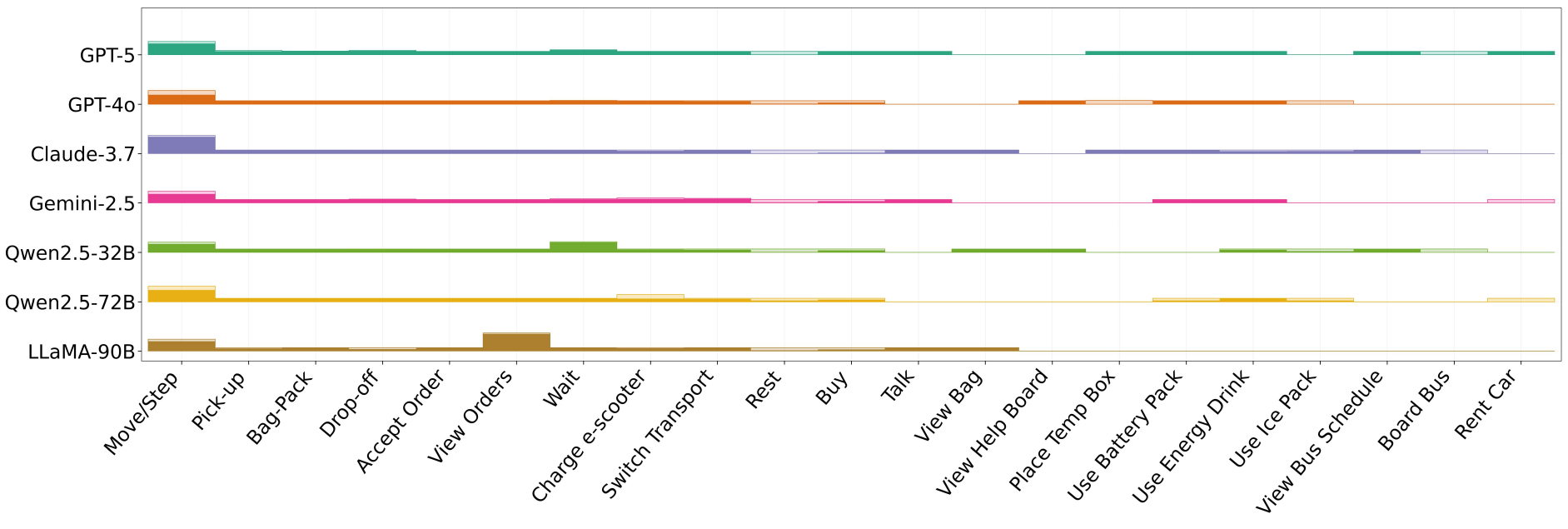}
    \vspace{-20pt}
    \caption{Action distributions of different models. For each model, the outer bars indicate the relative frequency of attempted actions, while the inner bars show the corresponding success rates.}
    \label{fig:actions}
    \vspace{-10pt}
\end{figure*}

\begin{figure}[!h]
    \centering
    \includegraphics[width=0.9\linewidth]{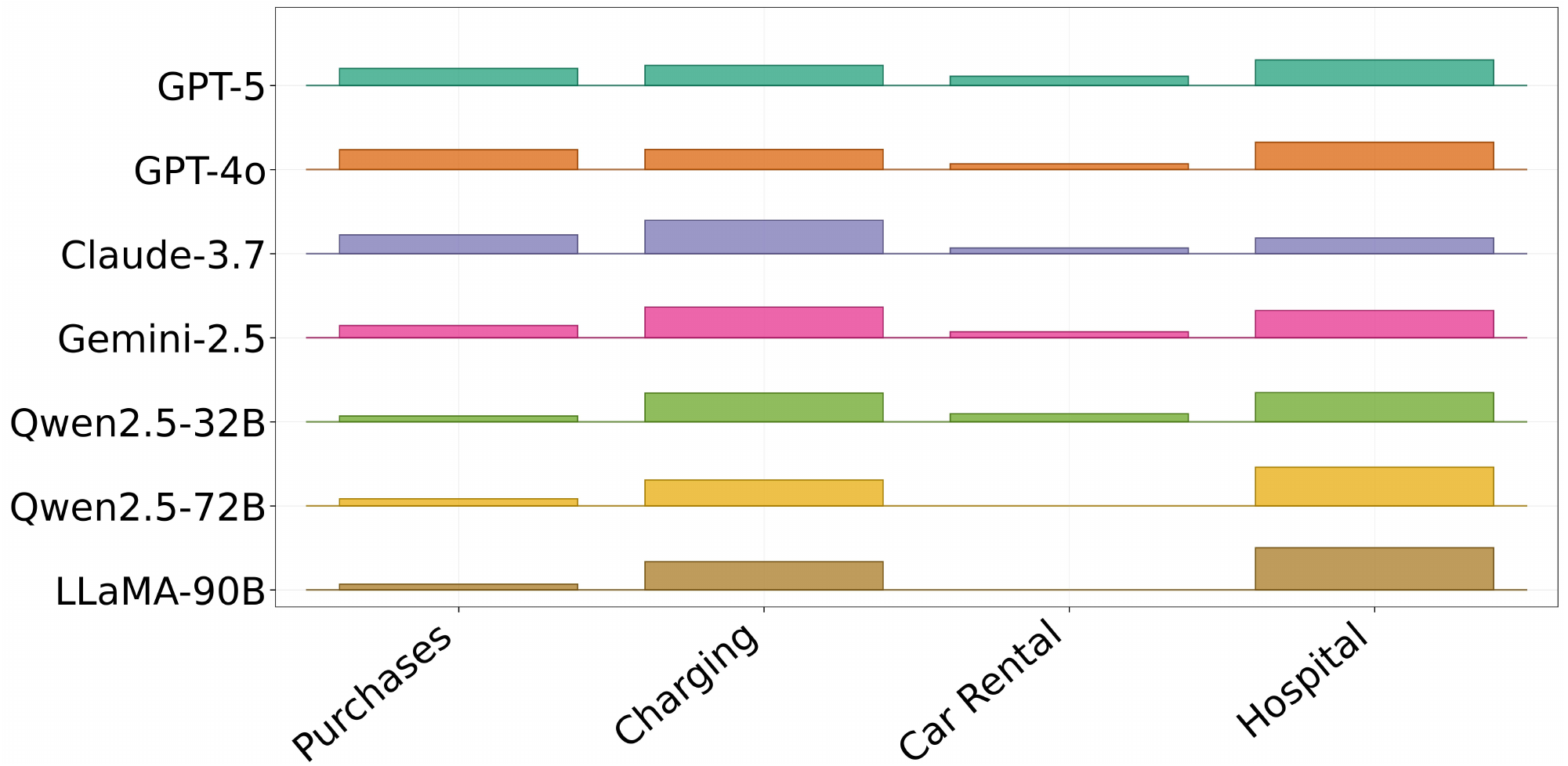}
    \caption{Expenditure distribution across models.}
    \label{fig:spend}
    \vspace{-10pt}
\end{figure}

\begin{figure}[!h]
    \centering
    \includegraphics[width=0.9\linewidth]{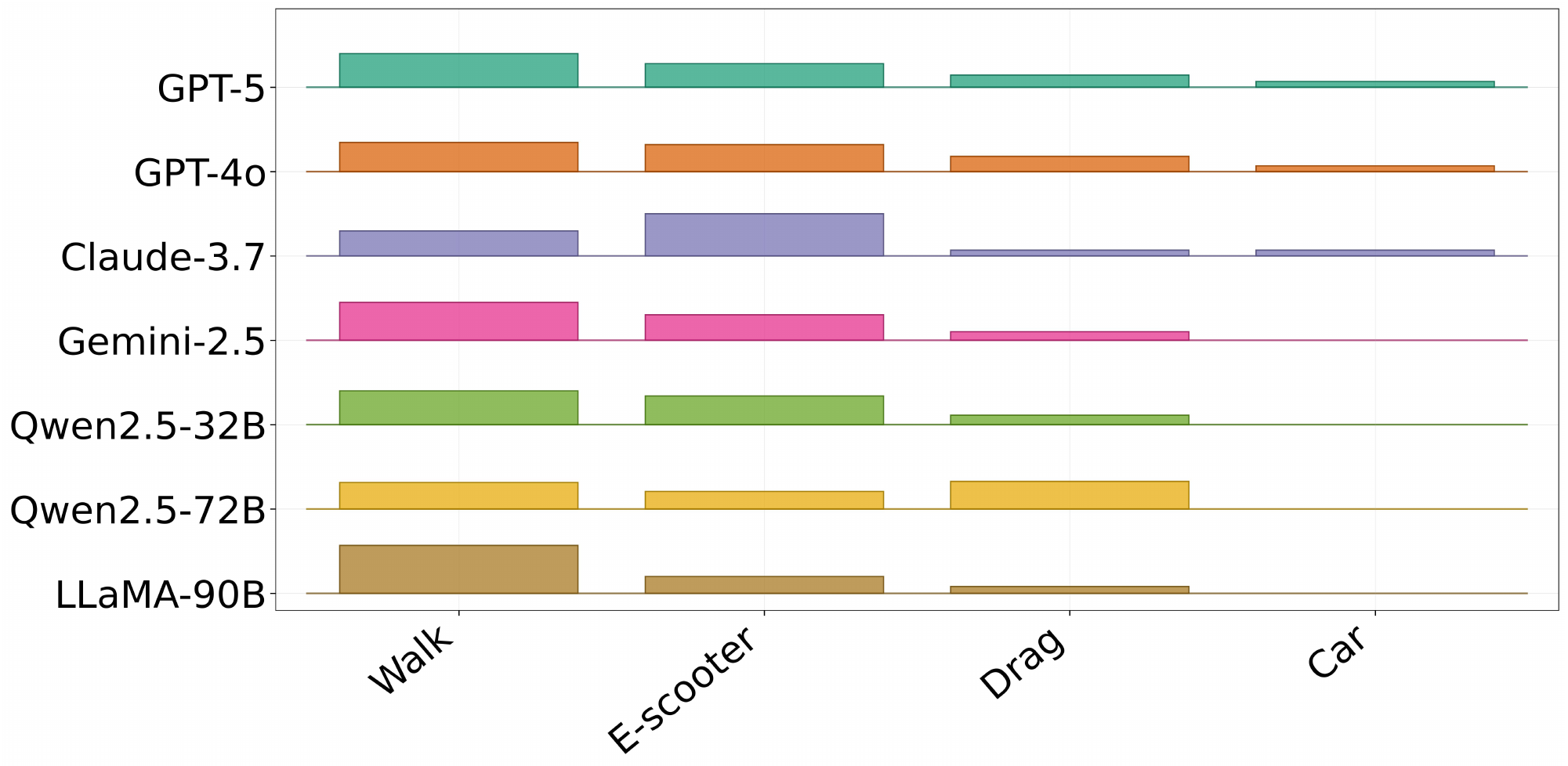}
    \vspace{-3pt}
    \caption{Transportation mode distribution across models.}
    \label{fig:transportation}
    \vspace{-10pt}
\end{figure}

In addition to the three examples of model planning styles shown in Figure~\ref{fig:planning_style}, we evaluate the behaviors of all models, with the remaining results presented in Figure~\ref{fig:planning_style_4}. We further analyze each model's action distribution, spending patterns, and transportation choices. As shown in Figure~\ref{fig:actions}, Stronger models such as GPT-5 and Claude-3.7-Sonnet exhibit broader action coverage and employ a richer set of strategies, such as renting cars or purchasing tools. In contrast, weaker open-source models such as LLaMA-3.2-90B-Vision-Ins primarily rely on simple pickup-and-delivery routines. These weaker models also end up spending more money on hospital rescues due to stamina depletion and often use less efficient transportation modes (e.g., walking or dragging scooters). Their spending patterns are summarized in Figure~\ref{fig:spend}, and their transportation preferences are illustrated in Figure~\ref{fig:transportation}.

\vspace{-5pt}

\subsection{Detailed Results for Context Engineering and Supervised Fine-tuning}
\label{appendix:context_sft_results}

We provide additional experimental results and analyses that complement the studies presented in Section~\ref{subsec:context_sft}, including more detailed metric breakdowns and illustrative case studies of model-generated summaries under context engineering.

\paragraph{Fine-grained Analysis.}
\label{appendix:context_sft_finegrained}

We further analyze model performance along three dimensions: high-level planning, resource management, and physical or environmental adaptation. As shown in Table~\ref{tab:ablation-noplan-full}, context engineering generally leads to higher on-time delivery rates, better time efficiency, and a larger active-time ratio, which allow the models to complete more orders and achieve higher earnings. However, the gains in resource management and environmental handling are less substantial. For the human-trajectory fine-tuning experiments, fine-tuning directly on raw actions results in noticeable declines across multiple capabilities. In contrast, fine-tuning on annotated trajectories produces significant improvements. In particular, time-efficiency scores even exceed those of large models such as GPT-5 and Claude-3.7-Sonnet, indicating that the model successfully learns the human strategy of handling multiple orders in parallel.

\begin{table*}[t]
\centering
\caption{
Fine-grained metrics for planning, resource usage, and physical/environmental behavior under context engineering and supervised fine-tuning. Green highlights improvements and red denotes regressions over the with-Plan baseline.
}
\vspace{-5pt}
\label{tab:ablation-noplan-full}
\scalebox{0.75}{
\renewcommand{\arraystretch}{1.05}
\begin{tabular}{l|cccc|ccc|ccc}
\toprule
\multirow{2}{*}{\textbf{Model}} &
\multicolumn{4}{c|}{\textbf{Planning}} &
\multicolumn{3}{c|}{\textbf{Resources}} &
\multicolumn{3}{c}{\textbf{Physical \& Env.}} \\
\cmidrule(lr){2-5} \cmidrule(lr){6-8} \cmidrule(l){9-11}
 & Order$\uparrow$ & OnTime$\uparrow$ & TimeEff$\uparrow$ & Active$\uparrow$
 & Stamina$\downarrow$ & Interrupts$\downarrow$ & Prevention$\uparrow$
 & Violations$\downarrow$ & Food$\uparrow$ & Cust$\uparrow$ \\
\midrule
GPT-5 (with Plan)             &
3.38 & 0.32 & 0.94 & 0.56 &
1.35 & 1.35 & 0.72 &
0.65 & 3.95 & 3.79 \\
GPT-5 (w/o Plan)              &
\cellcolor{red!12}3.24    & \cellcolor{red!12}0.25    & \cellcolor{red!12}0.45    & \cellcolor{red!12}0.48 &
\cellcolor{green!12}1.32  & \cellcolor{red!12}1.86    & \cellcolor{red!12}0.48 &
\cellcolor{red!12}0.75    & \cellcolor{red!12}3.35    & \cellcolor{red!12}3.20 \\
GPT-5 (with Plan + ACE)       &
\cellcolor{green!12}3.62   & \cellcolor{green!12}0.33  & \cellcolor{red!12}0.88      & \cellcolor{green!12}0.63 &
\cellcolor{red!12}1.66     & \cellcolor{red!12}2.50    & \cellcolor{red!12}0.62 &
\cellcolor{red!12}0.89     & \cellcolor{red!12}3.56    & \cellcolor{red!12}3.56 \\
GPT-5 (with Plan + DC)        &
\cellcolor{green!12}3.41   & \cellcolor{green!12}0.37  & \cellcolor{green!12}1.08     & \cellcolor{green!12}0.68 &
\cellcolor{green!12}1.29   & \cellcolor{red!12}2.96    & \cellcolor{green!12}0.79 &
\cellcolor{red!12}0.68     & \cellcolor{red!12}3.83    & \cellcolor{green!12}4.04 \\
\midrule
Claude-3.7-Sonnet (with Plan) &
3.46 & 0.41 & 0.92 & 0.59 &
0.78 & 0.64 & 0.77 &
0.62 & 3.80 & 3.72 \\
Claude-3.7-Sonnet (w/o Plan)  &
\cellcolor{red!12}3.28      & \cellcolor{red!12}0.37  & \cellcolor{red!12}0.58      & \cellcolor{red!12}0.54 &
\cellcolor{red!12}1.05      & \cellcolor{green!12}0.39  & \cellcolor{green!12}0.77 &
\cellcolor{red!12}0.78      & \cellcolor{green!12}3.88  & \cellcolor{green!12}3.76 \\
Claude-3.7-Sonnet (with Plan + ACE) &
\cellcolor{red!12}3.38   & \cellcolor{green!12}0.60    & \cellcolor{green!12}0.96   & \cellcolor{green!12}0.82 &
\cellcolor{red!12}0.79   & \cellcolor{green!12}0.50    & \cellcolor{green!12}0.91 &
\cellcolor{red!12}0.70   & \cellcolor{green!12}4.00    & \cellcolor{green!12}4.30 \\
Claude-3.7-Sonnet (with Plan + DC) &
\cellcolor{red!12}3.41   & \cellcolor{green!12}0.52    & \cellcolor{green!12}1.06   & \cellcolor{green!12}0.77 &
\cellcolor{red!12}1.22   & \cellcolor{red!12}1.06      & \cellcolor{red!12}0.54 &
\cellcolor{red!12}0.72   & \cellcolor{green!12}3.92    & \cellcolor{green!12}4.16 \\
\midrule
Qwen2.5-VL-72B (with Plan)    &
3.12 & 0.17 & 0.40 & 0.53 &
1.38 & 1.50 & 0.53 &
0.70 & 4.11 & 3.73 \\
Qwen2.5-VL-72B (w/o Plan)     &
\cellcolor{red!12}3.07      & \cellcolor{green!12}0.21  & \cellcolor{red!12}0.38      & \cellcolor{red!12}0.51 &
\cellcolor{red!12}1.42      & \cellcolor{red!12}2.13    & \cellcolor{red!12}0.24 &
\cellcolor{red!12}0.75      & \cellcolor{red!12}3.61    & \cellcolor{red!12}3.35 \\
Qwen2.5-VL-72B (with Plan + ACE) &
\cellcolor{red!12}2.97  & \cellcolor{red!12}0.14      & \cellcolor{green!12}0.88   & \cellcolor{green!12}0.63 &
\cellcolor{red!12}1.76  & \cellcolor{red!12}3.00      & \cellcolor{red!12}0.40 &
\cellcolor{red!12}1.00  & \cellcolor{red!12}3.80      & \cellcolor{red!12}3.40 \\
Qwen2.5-VL-72B (with Plan + DC) &
\cellcolor{green!12}3.49  & \cellcolor{green!12}0.36      & \cellcolor{green!12}0.59   & \cellcolor{green!12}0.72 &
\cellcolor{green!12}0.98  & \cellcolor{green!12}1.26      & \cellcolor{red!12}0.44 &
\cellcolor{green!12}0.62  & \cellcolor{green!12}4.16      & \cellcolor{green!12}4.03 \\
\midrule
LLaVA-OneVision-8B (original)       &
3.22 & 0.05 & 0.15 & 0.50 &
2.32 & 2.49 & 0.16 &
0.74 & 3.67 & 3.52 \\
LLaVA-OneVision-8B (human-ft)       &
\cellcolor{red!12}3.05     & \cellcolor{green!12}0.06    & \cellcolor{green!12}0.72  & \cellcolor{red!12}0.38 &
\cellcolor{red!12}2.49     & \cellcolor{red!12}2.99      & \cellcolor{red!12}0.14    &
\cellcolor{red!12}0.82     & \cellcolor{red!12}3.63      & \cellcolor{red!12}3.04 \\
LLaVA-OneVision-8B (annotated-ft)   &
\cellcolor{green!12}3.36   & \cellcolor{green!12}0.16    & \cellcolor{green!12}1.51  & \cellcolor{green!12}0.88 &
\cellcolor{green!12}0.64   & \cellcolor{green!12}2.38    & \cellcolor{green!12}0.47  &
\cellcolor{green!12}0.58   & \cellcolor{green!12}4.02    & \cellcolor{green!12}3.96 \\
\bottomrule
\end{tabular}}
\vspace{-8pt}
\end{table*}

\paragraph{Context Engineering Case Study.}
\label{appendix:contextstudy}

\begin{figure*}[!h]
    \centering
    \includegraphics[width=0.9\linewidth]{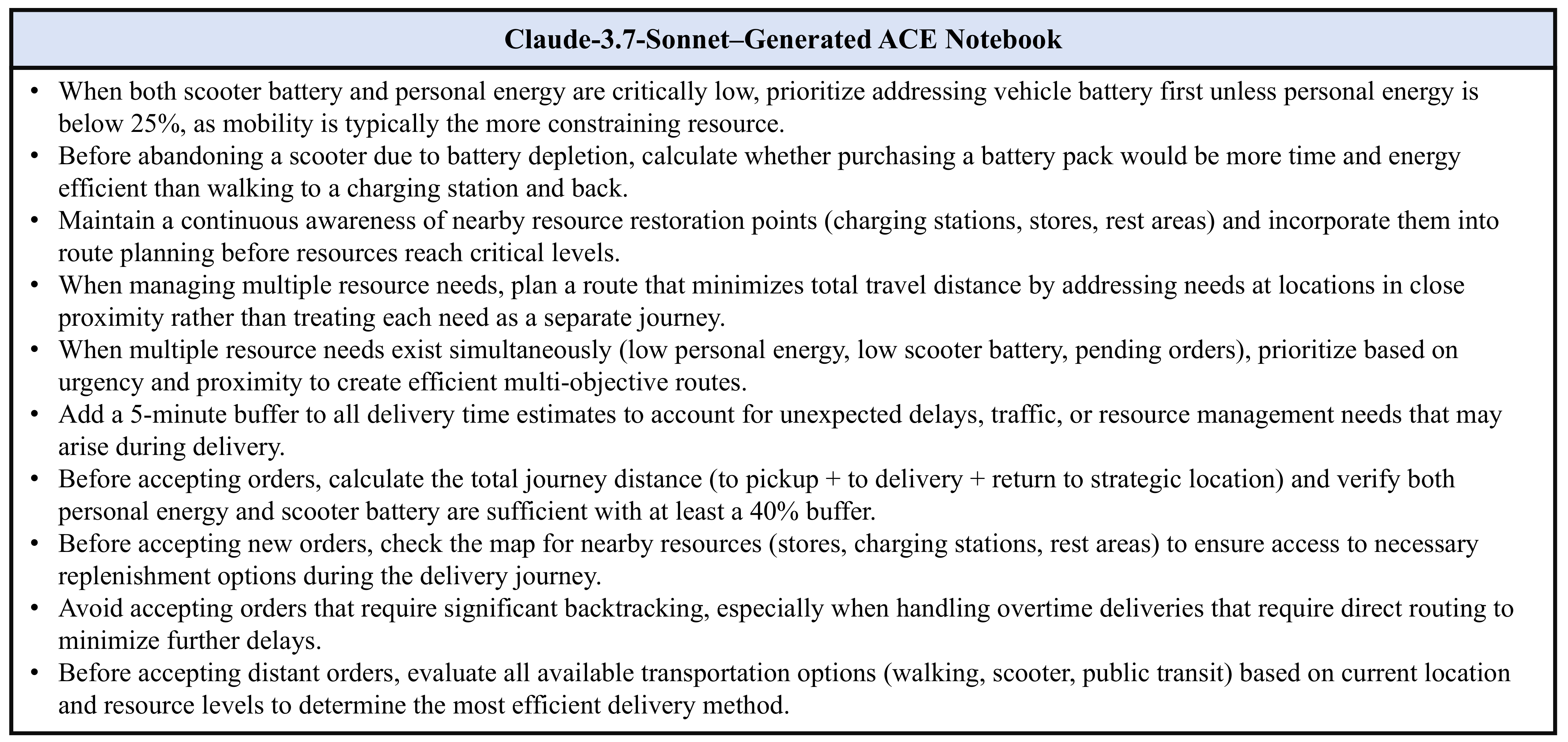}
    \vspace{-8pt}
    \caption{Example ACE notebook generated by Claude-3.7-Sonnet.}
    \label{fig:ACE_Claude}
    \vspace{-5pt}
\end{figure*}

\begin{figure*}[!h]
    \centering
    \includegraphics[width=0.9\linewidth]{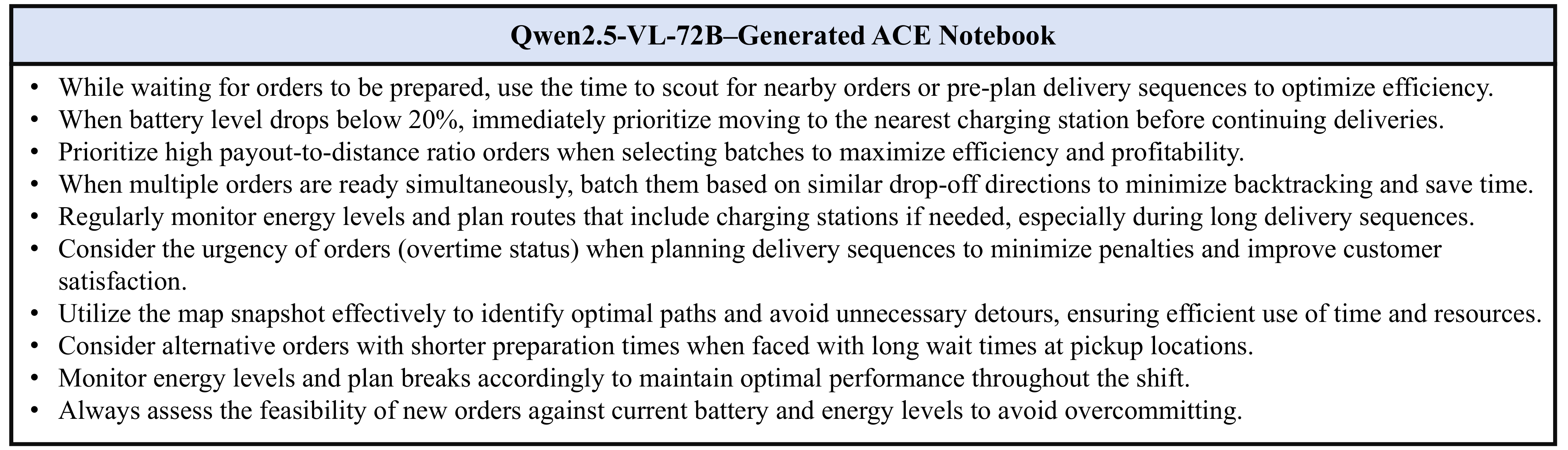}
    \vspace{-8pt}
    \caption{Example ACE notebook generated by Qwen2.5-VL-72B.}
    \label{fig:ACE_Qwen}
    \vspace{-5pt}
\end{figure*}

We present example notebooks generated by Claude-3.7-Sonnet and Qwen2.5-VL-72B under Agentic Context Engineering (ACE). In this setting, each model autonomously summarizes patterns from its past trajectories and maintains these summaries as persistent memory to guide future deliveries. For each model, we select the ten highest-quality examples, shown in Figure~\ref{fig:ACE_Claude} and Figure~\ref{fig:ACE_Qwen}. Both models extract principles covering multiple aspects of delivery, including time management and resource planning, and their summaries closely align with the underlying task rules. In comparison, Claude-3.7-Sonnet produces more detailed and actionable guidelines, which in turn contributes to its larger performance improvement when ACE is applied.

\subsection{Ablation Studies}

\label{sec:exploratory}

\noindent \textbf{Planning Ablation.}
We further analyze the results reported in Table~\ref{tab:ablation-noplan}, which compare models that perform explicit plan-and-execute reasoning with models that directly output a single action. For GPT-5 and Qwen2.5, planning consistently improves most capability metrics and leads to higher net profit. In contrast, Claude-3.7-Sonnet earns more when planning is enabled, but its net profit decreases because of increased expenses. These additional costs mainly arise from overplanning, such as repeatedly recharging the e-scooter when the battery level is already sufficient or purchasing items that are not immediately necessary.

\vspace{5pt}
\noindent \textbf{Waypoint Ablation.}
We evaluate whether VLM agents can navigate without privileged spatial priors. We remove preset waypoints and restrict them to step-by-step navigation using only low-level actions (STEP\_FORWARD, TURN\_AROUND) with egocentric observations. Agents fail to complete even a single order under this setting, indicating that current models struggle to translate visual understanding into embodied navigation. Explicit spatial coordinates remain a dependency for these models.

\subsection{Variance and Stability Analysis}
\label{appendix:variance}

We further evaluate the stability of model performance under repeated runs. For both Gemini-2.5-Flash and Qwen2.5-VL-72B-Ins, we conduct three experimental groups, each following the same setup as the main experiment and consisting of eight independent runs along with their averaged results. As shown in Table~\ref{tab:overall-metrics-vertical}, overall, both models exhibit low variance across runs, demonstrating stable and reliable performance under identical conditions.

\begin{table}[!h]
\caption{Mean values and run-to-run variability for Gemini-2.5-Flash and Qwen2.5-VL-72B-Ins.}
\label{tab:overall-metrics-vertical}
\centering
\scalebox{0.9}{
\begin{tabular}{l|c|c}
\toprule
\textbf{Metric} & \textbf{Gemini-2.5-Flash} & \textbf{Qwen2.5-VL-72B-Ins} \\
\midrule
$\bar{P}$             & \$28.46 $\pm$ 2.52 & \$5.96 $\pm$ 2.82 \\
$E$ & \$37.55 $\pm$ 2.11 & \$13.28 $\pm$ 2.90 \\
$C$                   & -\$9.09 $\pm$ 1.32 & -\$7.32 $\pm$ 0.96 \\
Order      & 3.32 $\pm$ 0.11 & 3.07 $\pm$ 0.09 \\
OnTime      & 0.30 $\pm$ 0.07 & 0.18 $\pm$ 0.05 \\
TimeEff      & 0.88 $\pm$ 0.08 & 0.45 $\pm$ 0.04 \\
Active      & 0.52 $\pm$ 0.04 & 0.50 $\pm$ 0.05 \\
Stamina      & 1.03 $\pm$ 0.06 & 1.42 $\pm$ 0.09 \\
Interrupts      & 1.79 $\pm$ 0.05 & 1.57 $\pm$ 0.10 \\
Prevention      & 0.78 $\pm$ 0.06 & 0.55 $\pm$ 0.08 \\
Violations                   & 0.70 $\pm$ 0.11 & 0.68 $\pm$ 0.14 \\
Food     & 4.08 $\pm$ 0.11 & 4.01 $\pm$ 0.17 \\
Cust     & 3.77 $\pm$ 0.20 & 3.62 $\pm$ 0.25 \\
\bottomrule
\end{tabular}}
\end{table}

\end{document}